\documentclass[lettersize,journal]{IEEEtran}
\usepackage{amsmath,amsfonts}
\usepackage{algorithmic}
\usepackage{algorithm}
\usepackage{array}
\usepackage[caption=false,font=normalsize,labelfont=sf,textfont=sf]{subfig}
\usepackage{textcomp}
\usepackage{stfloats}
\usepackage{url}
\usepackage{verbatim}
\usepackage{pifont}
\usepackage{cite}
\usepackage{wrapfig}
\usepackage{colortbl}
\usepackage{tabularx}
\usepackage{multirow}
\usepackage{arydshln}
\usepackage{float}
\usepackage{booktabs}
\usepackage{graphicx}
\usepackage{hyperref}
\usepackage{amssymb}
\usepackage{mathrsfs}
\usepackage{amsmath}

\definecolor{mygray}{gray}{.9}
\definecolor{linkcolor}{RGB}{255,0,0}
\definecolor{urlcolor}{RGB}{255,105,180}
\definecolor{citecolor}{RGB}{66,168,235}

\begin{document}
\definecolor{HBSky}{RGB}{84,151,244}
\definecolor{HBGreen}{RGB}{97,143,0}
\definecolor{HBOrange}{RGB}{248,149,91}
\definecolor{HBRed}{RGB}{255,0,0}
\newcommand{\Rone}[1]{\textcolor{HBSky}{#1}}

\newcommand{\Rtwo}[1]{\textcolor{HBGreen}{#1}}
\newcommand{\orange}[1]{\textcolor{HBOrange}{#1}}
\newcommand{\Rred}[1]{\textcolor{HBRed}{#1}}

\newcommand{\blue}[1]{\textcolor{HBSky}{#1}}

\title{DeGMix: Efficient Multi-Task Dense Prediction with Deformable and Gating Mixer}
\author{Yangyang Xu, \and Yibo Yang, \and Bernard Ghanem, \textit{Member, IEEE}, \and Lefei Zhang,  \textit{Senior
Member, IEEE}, Bo Du, \textit{Senior Member, IEEE}, and Jun Zhu, \textit{Fellow, IEEE}
\thanks{\textit{Yangyang Xu and Jun Zhu are with the Department of Computer Science \& Technology, Institute for AI, Tsinghua University, Beijing, 100084, China. E-mail: {yangyangxu}@mail.tsinghua.edu.cn}; dcszj@tsinghua.edu.cn.}
\thanks{\textit{Lefei Zhang and Bo Du are with the School of Computer Science, Wuhan University, Wuhan 430072, China. E-mail: \{zhanglefei,dubo\}@whu.edu.cn}.}
\thanks{\textit{Yibo Yang and Bernard Ghanem are with the King Abdullah University of Science and Technology, Thuwal, Mecca Province, 23955-6900, Saudi Arabia. E-mail:\{yibo.yang,Bernard.Ghanem\}@kaust.edu.sa}.}
\thanks{\textit{Manuscript received April xx, 2025; revised xx xx, 202x.}}
\thanks{\textit{Corresponding Author: Lefei Zhang.}}

}

\markboth{Journal of \LaTeX\ Class Files,~Vol.~14, No.~8, August~2021}%
{Shell \MakeLowercase{\textit{et al.}}: A Sample Article Using IEEEtran.cls for IEEE Journals}


\maketitle

\begin{abstract}
Convolution neural networks (CNNs) and Transformers have their own advantages and both have been widely used for dense prediction in multi-task learning (MTL).
CNNs excel at capturing local spatial patterns, whereas Transformers are effective at modeling long-range dependencies. However, existing approaches typically employ them independently, leaving the potential of their complementary strengths underexploited.
In this work, we present \textbf{DeGMix} (\textbf{De}formable and \textbf{G}ating \textbf{Mix}er), an efficient encoder–decoder framework that unifies convolution and attention mechanisms for MTL. 
DeGMix integrates a \emph{deformable mixer encoder} and a \emph{task-aware gating Transformer decoder} to achieve robust and parameter-efficient learning.
First, the deformable mixer encoder contains two types of operators: the channel-aware mixing operator leveraged to allow communication among different channels, and the spatial-aware deformable operator with deformable convolution applied to efficiently sample more informative spatial locations.
By simply stacking the operators, we obtain the deformable mixer encoder, which effectively captures significant deformable features.
Second, the task-aware gating transformer decoder is used to perform task-specific predictions, in which task interaction block integrated with self-attention is applied to capture task interaction features, and the task query block integrated with gating attention is leveraged to dynamically select the corresponding task-specific features.
Furthermore, the results of the experiment demonstrate that the proposed DeGMix uses fewer GFLOPs and significantly outperforms current Transformer-based and CNN-based competitive models on a variety of metrics on three dense prediction datasets (\textit{i.e.,} NYUD-v2, PASCAL-Context, and Cityscapes).
For example, using Swin-L as the backbone, our method achieves 57.55 mIoU in segmentation on NYUD-v2, outperforming the best existing method by +5.2 mIoU.
Our fine-tuning experiments on the PASCAL-Context show that DeGMix achieves higher accuracy on downstream tasks compared to the MTL-fully fine-tuning model while reducing the number of trainable parameters by 9.39$\times$.
Our code and models are publicly available at {\url{https://github.com/yangyangxu0/DeGMix}}. 
\end{abstract}


\section{Introduction}\label{sec:intro}
The human visual system possesses remarkable capabilities, enabling it to perform a multitude of tasks within a single visual scene, including classification, segmentation, and recognition.
Consequently, multi-task learning (MTL) has become a significant area of research in computer vision.
Our objective is focused on developing an advanced vision model capable of simultaneously performing multiple tasks across diverse visual scenarios, with a strong emphasis on efficiency.
The core challenge lies in designing an efficient architecture that can effectively capture both task-specific and shared representations across tasks like semantic segmentation, human parts segmentation, depth estimation, boundary detection, saliency estimation, and normal estimation, all while maintaining high performance and computational efficiency.

Recently, existing works~\cite{Mti-net_2020,multitask_mtst_2021,atrc_2021,mqtrans_xy,transfmult2022,InvPT_2022} have achieved significant advancements through CNN and Transformer technologies for MTL of dense prediction. 
CNN-based MTL models excel at local representation learning and have been carefully engineered for cross-task feature sharing; $e.g.,$ Cross-stitch~\cite{Cross-Stitch_2016} mixes shared and task-specific activations, while others~\cite{atrc_2021,Mti-net_2020} leverage distillation and enlarged receptive fields to strengthen cross-task information flow.
Transformer-based MTL models~\cite{transfmult2022,mqtrans_xy,xu2022mtformer} utilize the efficient self-attention mechanism~\cite{attent2017}, which facilitates efficient modeling of global interactions and task interactions.
These approaches demonstrate the growing potential of CNNs and Transformers in enhancing the performance of multi-task dense prediction.

We, in particular, identify three primary challenges in MTL. 
First, although CNN-based MTL models have been proven effective for MTL methods~\cite{NDDR-CNN_2019,Mti-net_2020,multitask_mtst_2021}, which better capture the multiple task context in a local field, they suffer from a lack of global modeling and task interaction.
These CNN-based MTL methods usually focus on modeling locality naturally, thus lacking the capability to model long-range dependencies. 
Recently, there have been a few attempts~\cite{atrc_2021} to introduce the global and local context through distillation using scaled dot-product attention.
However, these distillation pipelines often struggle to fully address global dependencies and tend to introduce unnecessary complexity.

Second, existing Transformer-based models~\cite{transfmult2022,InvPT_2022} excel at modeling the global dependencies across tasks and enhancing task interactions in MTL.
However, these models rely on standard self-attention mechanisms, which overlook the task-specific local information and introduce huge computational costs.
Specifically, it only focuses on capturing long-range dependencies without task-aware local information during self-attention processing.
This results in significant computational demands, particularly when dealing with high-resolution inputs, as the calculations involving queries, keys, and values become increasingly intensive.

Third, although several methods\cite{xu2023demt} combine CNNs and Transformers to capture both local and global features, they struggle to effectively select task-relevant features in the corresponding task. 
This limitation can undermine the overall performance of the model and its ability to adapt effectively across a range of diverse tasks, ultimately affecting its versatility and generalization capabilities.

Recent works~\cite{swin,vitadaptor2022} demonstrate that locality is a reasonable compensation for global dependency.
However, the potential benefits of integrating CNN and Transformer in MTL have yet to be fully explored.
These challenges motivate us to seek a principled combination of CNN and Transformer architectures that can inherit both local sensitivity and global reasoning for multi-task dense prediction in MTL.
To address the challenges, we propose the deformable and gating mixer (DeGMix) model: a simple and effective approach to multi-task dense prediction.
DeGMix leverages the strengths of deformable CNNs and query-based Transformers, enhanced with spatial gating mechanisms, to deliver superior performance. 
Instead of a naive combination of CNN and Transformer, we introduce an innovative design comprising a deformable mixer encoder and a task-aware gating Transformer decoder, ensuring a seamless integration of their complementary capabilities.

Specifically, the deformable mixer encoder of our DeGMix consists of channel-aware mixing convolution and spatial-aware deformable convolution.
Motivated by the success of deformable convolutional networks~\cite{defcnn_2019} in vision tasks,
our deformable mixer encoder learns different deformed features for each task based on more efficient sampling spatial locations and channel location mixing ($i.e.,$ deformed feature).
It learns multiple deformed features that highlight more informative regions with respect to the different tasks. 
The task-aware gating transformer decoder of our DeGMix consists of task interaction block and task query gating blocks.
In the task-aware gating transformer decoder, the multiple deformed features are fused and fed into our task interaction block. 
We use the fused feature to generate task-interacted features via a multi-head self-attention for model task interactions.
Current Transformer-based studies for MTL use regular query tokens. 
In contrast, we use deformed features as query tokens via attention with gating, which increases the task awareness for the final representation output by the Task-aware Gating Transformer Decoder.
Thus, we use deformed features directly as \textit{query} tokens to focus on the task awareness of each task.
We expect the set of candidate \textit{key/value} to be from task-interacted features.
Then, our task query gating block uses the deformed features and task-interacted features as input and generates the dynamically selected task awareness features.

In this way, our deformable mixer encoder adaptively selects informative spatial regions as deformed features to alleviate the lack of global modeling in CNN.
The task-aware gating transformer decoder performs the task interactions by self-attention and enhances task awareness via a query-based Transformer with a spatial gating.
By integrating the complementary advantages of both modules, the framework delivers deformable and comprehensive representations from local and global perspectives, while maintaining efficiency and emphasizing task-specific relevance.

\begin{figure}[!t]
\centering
  \includegraphics[width=0.5\textwidth]{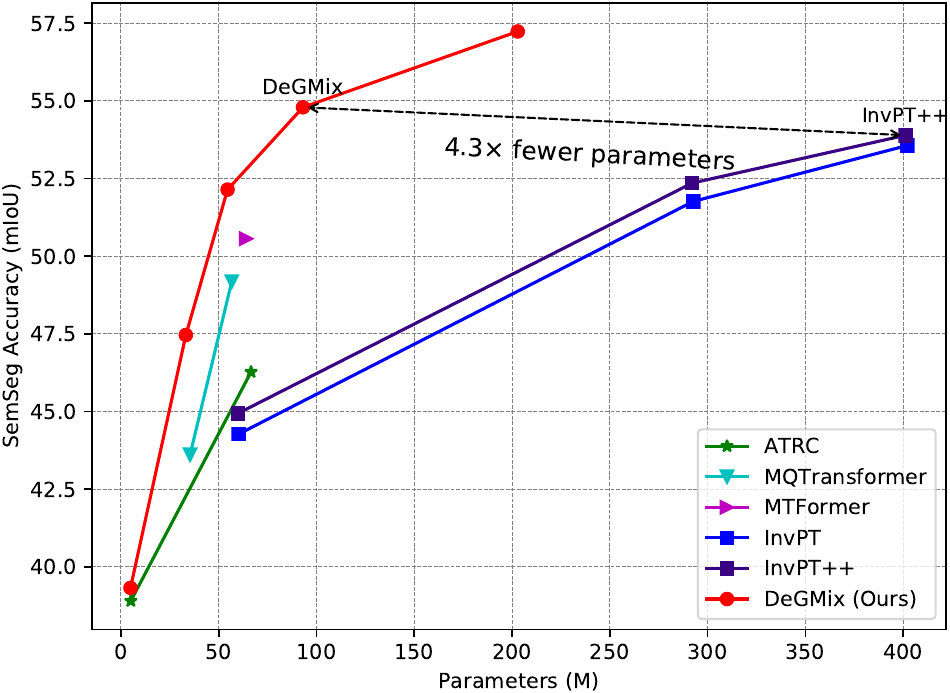}
  \caption{Accuracy-to-Parameter trade-off on NYUD-v2 dataset. Our DeGMix model can outperform existing MTL methods by a large margin while maintaining an optimal balance between accuracy and model parameters. }
  \label{fig:params}
\end{figure}

We conduct extensive experiments on three dense prediction datasets to evaluate the DeGMix.
Our DeGMix approach consistently outperforms the state-of-the-art methods on a variety of metrics (see Figure~\ref{fig:params}).
We perform ablation studies to validate the effectiveness of DeGMix, highlighting its key strengths and unique properties. 
We also fine-tune\cite{VPTshallow2022} DeGMix for a parameter-efficient MTL approach.
The main contributions of this paper are summarized as follows:

1) We propose a simple and effective DeGMix method for MTL of dense prediction via combining both merits of CNN and Transformer. 
Most importantly, our method addresses the limitations of CNN-based models by incorporating global modeling capabilities, while simultaneously overcoming the task-awareness shortcomings of Transformer-based models.

2) We introduce an efficient multi-task dense prediction with deformable and gating mixer (DeGMix) model, which consists of the deformable mixer encoder (Section~\ref{subsec:enc}) and task-aware gating transformer decoder (Section~\ref{subsec:dec}). 
DeGMix can adaptively control task-specific feature learning by capturing deformable features using deformable convolution with spatial gating in the deformable mixer encoder, as well as task-aware features through query-based attention with spatial gating in the task-aware gating transformer decoder.

3) A novel shared spatial gating layer is introduced across all tasks to perform dynamic feature selection.
This gating mechanism effectively highlights informative features and ensures that only task-relevant information is propagated through both the encoder and the decoder, leading to high-quality, task-specific outputs, enhancing the model's ability to effectively learn and distinguish between different tasks.

4) Extensive experiments are conducted in NYUD-v2~\cite{NYUD2012}, PASCAL-Context~\cite{pascal2014}, and Cityscapes~\cite{2016cityscapes}. 
DeGMix demonstrates superior accuracy in downstream tasks compared to fully fine-tuning MTL model while training significantly fewer parameters.
The visualization results show the efficacy of our model. DeGMix's strong performance on MTL can demonstrate the benefits of combining the deformable CNN and query-based Transformer. As shown in Figure~\ref{fig:params}, it outperforms all state-of-the-art methods by a large margin on SemSeg.

\section{Related Work}
\label{sec:relate}

\subsection{Multi-Task Learning (MTL)}
Multi-task learning (MTL) has become an increasingly popular area of research in recent years, with many works proposing new methods and architectures to tackle the challenges of learning multiple tasks simultaneously. 
MTL has undergone significant evolution with the advancements in CNNs and Vision Transformers.
MTL tasks are primarily distributed in two aspects: the development of innovative model architectures~\cite{atrc_2021} and the optimization of task loss weighting~\cite{transfmult2022}. 
In the vision domain, the core idea of MTL is to use a single model to predict semantic segmentation, human parts segmentation, depth, surface normal, boundary, $etc.,$ which is an interesting topic.
Recent works propose novel architectures that learn shared representation to capture both task-specific and shared information, enhancing overall performance. 
MuST~\cite{multitask_mtst_2021} model uses the knowledge in independent specialized teacher models to train a general model for creating general visual representations.
Several recent MTL frameworks follow different technologies:~\cite{NDDR-CNN_2019} and \cite{Mti-net_2020} are CNN-based MTL models,~\cite{atrc_2021} is Neural Architecture Search-based model and~\cite{transfmult2022,mqtrans_xy,xu2022mtformer,InvPT_2022} are Transformer-based models.

\cite{MTL_survey_2021} states that the MTL structures in vision tasks can be summarized into two categories: encoder-focused and decoder-focused architectures.
Encoder-focused approaches~\cite{encoder_base_2018,encoder_base2_2018} employ a shared encoder to extract general visual representations, which are then processed by task-specific heads for individual predictions.
Decoder-focused works~\cite{NDDR-CNN_2019,PAP-Net_2019,atrc_2021,invpt++} use a shared backbone network to extract a shared feature for each task. Then, the designed task-specific modules perform the interactions to capture valuable information from the cross-task.
Representative examples include MTI-Net~\cite{Mti-net_2020}, which models task interactions at multiple scales through feature propagation and aggregation, and ATRC~\cite{atrc_2021}, which adaptively selects the most informative attention context for task refinement.

With the success of Vision Transformers~\cite{ViT2021}, recent advances in computational modeling have significantly improved MTL performance~\cite{xu2022mtformer,InvPT_2022,mqtrans_xy,invpt++}.
MTFormer~\cite{xu2022mtformer} and InvPT~\cite{InvPT_2022} enhance dense prediction through self-attention mechanisms built on strong Transformer backbones.
In contrast, MQTransformer~\cite{mqtrans_xy} introduces multi-task queries to explicitly model task interactions and transfer task-specific knowledge.
However, most Transformer-based MTL methods still emphasize shared representations, limiting their ability to disentangle task-specific features.

\subsection{Deformable CNNs and Vision Transformers.}
\textbf{Deformable CNNs.} 
CNNs have achieved milestone contributions in many domains.
Deformable ConvNets~\cite{deformablecnn2017,defcnn_2019} harness the enriched modeling capability of CNNs via deformable convolution and deformable spatial locations.
A deformable convolution dynamically attends to flexible spatial locations conditioned on learned offsets, enabling more expressive feature modeling.
Inspired by this idea, deformable attention~\cite{deformableDetr} extends the concept to Transformers by attending to a sparse set of informative sampling points, efficiently capturing local-to-global dependencies.
CNNs are widely used for MTL, and recent works have proposed various CNN-based architectures to learn shared features across tasks.
UberNet~\cite{kokkinos2017cnn_MTL} is proposed to train in an end-to-end manner a unified CNN architecture to solve multiple vision tasks.
NDDR-CNN~\cite{NDDR-CNN_2019} introduces a CNN structure, neural discriminative dimensionality reduction, for multi-task learning, which enables automatic feature fusing at every layer and is formulated by combining existing CNN components in a novel way with clear mathematical interpretability.
In addition, MTI-Net~\cite{Mti-net_2020} proposes a MTL method, which models task interactions at multiple scales through a multi-scale multi-modal distillation unit, feature propagation module, and feature aggregation unit to produce refined task features and improve performance on dense prediction tasks. 
Previous CNN-based MTL methods utilize plain convolution to capture the task's local information from the input feature. 
However, conventional CNN-based MTL models mainly rely on local receptive fields and insufficiently exploit global context.
In contrast, our approach replaces the self-attention variables (query, key, and value) in Transformers with deformable feature tensors, integrating the flexibility of deformable convolutions with the global reasoning ability of Transformers.

\noindent
\textbf{Vision Transformers.}
Transformers and attention mechanisms~\cite{attent2017} have shown great success in natural language processing and have been rapidly extended to computer vision tasks such as detection, classification, and segmentation.
Vision Transformers (ViT)~\cite{ViT2021} pioneered the adaptation of self-attention to image analysis by representing an image as a sequence of visual patches, achieving performance comparable to or surpassing modern CNNs.
Subsequent works~\cite{swin,Ranftl2021} have been introduced by improving the attention mechanism for dense prediction tasks.
Unlike the original designs in~\cite{attent2017,ViT2021}, some vision Transformers usually employ a hierarchical architecture and replace the global self-attention among all patches to local self-attention to develop efficient attention mechanisms. 
These efficient attention mechanisms include windowed attention~\cite{dong2022cswin,swin}, global tokens~\cite{jaegle2021globalatten}, gated attention~\cite{yang2022focalgating} and deformable attention~\cite{deformableDetr}. 
Recently, these works are also extended to the MTL domain to learn good representations for multiple tasks of dense predictions. 
MulT~\cite{transfmult2022} uses a shared attention mechanism to model task dependencies and perform MTL for dense prediction.
MTFormer~\cite{xu2022mtformer} designs the shared transformer encoder and shared transformer decoder, and task-specific self-supervised cross-task branches are introduced to generate the final outputs, thereby enhancing the MTL performance. 
InvPT~\cite{InvPT_2022} and InvPT++\cite{invpt++} employ an inverted pyramid multi-task transformer to learn long-range interaction via plain self-attention in both spatial and all-task contexts on the multi-task features with a gradually increased spatial resolution for dense prediction.
In the DeMT~\cite{xu2023demt} method, convolution and attention are combined to perform muli-task dense prediction.
In contrast, our framework introduces two key innovations: \textbf{First}, we introduce a novel simplified gating mechanism in the deformable mixer encoder and task-aware gating transformer decoder to selectively gather contexts for each query token based on its content. 
\textbf{Second}, we design a shared spatial gating layer that applies spatial gating to each task-specific feature within the task query gating block the most informative regions.
In addition, we use the query-based transformer with task shared gating block approach for modeling and leverage deformed features as queries in transformer calculations to enhance task-relevant features. 
These queries can naturally disentangle the task-specific feature from the fused feature.
By integrating deformable convolutional flexibility with Transformer-based global reasoning, our method achieves state-of-the-art results on NYUD-v2~\cite{NYUD2012}, PASCAL-Context~\cite{pascal2014}, and Cityscapes~\cite{2016cityscapes}.

\section{Approach}
\label{sec:method}

\subsection{Overall Architecture}

DeGMix adopts a non-shared encoder–decoder architecture.
Specifically, a deformable mixer encoder extracts task-specific spatial representations, while the task interaction and task query gating blocks decode task-aware features through self-attention with gating mechanisms.
The following section presents the multi-task learning objectives.

\begin{figure*}[!t]
\centering
  \includegraphics[width=0.85\textwidth]{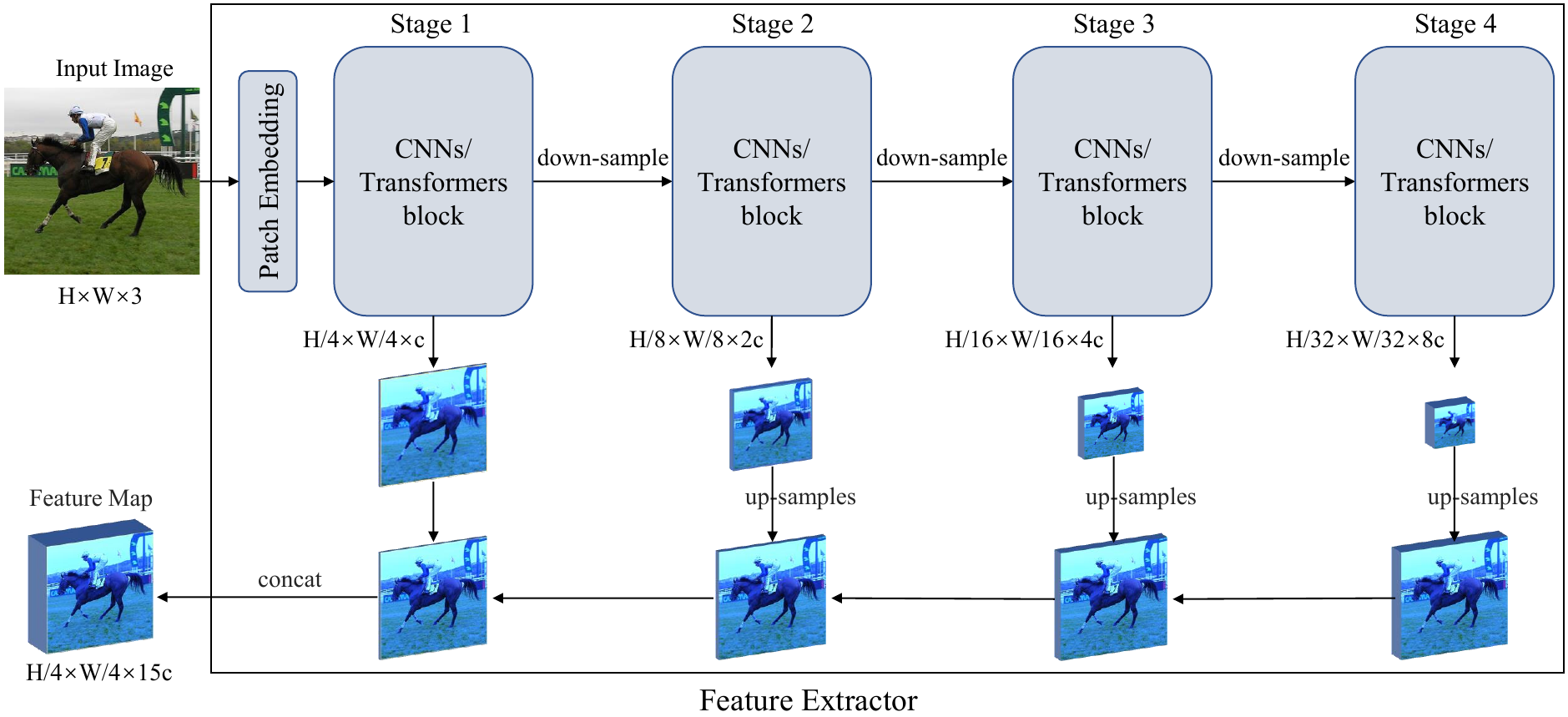}
  \caption{Illustration of our feature extractor. Our feature extractor can be compatible with three kinds of backbone networks (\textit{i.e.,} HRNet, Swin-Transformer (Swin) and Vision Transformer (ViT)). Note that when ViT is used as backbone, we select four block-specific features. We conduct the scale number ablation study in Table~\ref{tab:scales}. In this figure, we set the feature map channel $15c=C$.}
  \label{fig:extractor}
\end{figure*}

\begin{figure*}[ht]
\centering
  \includegraphics[width=0.96\textwidth]{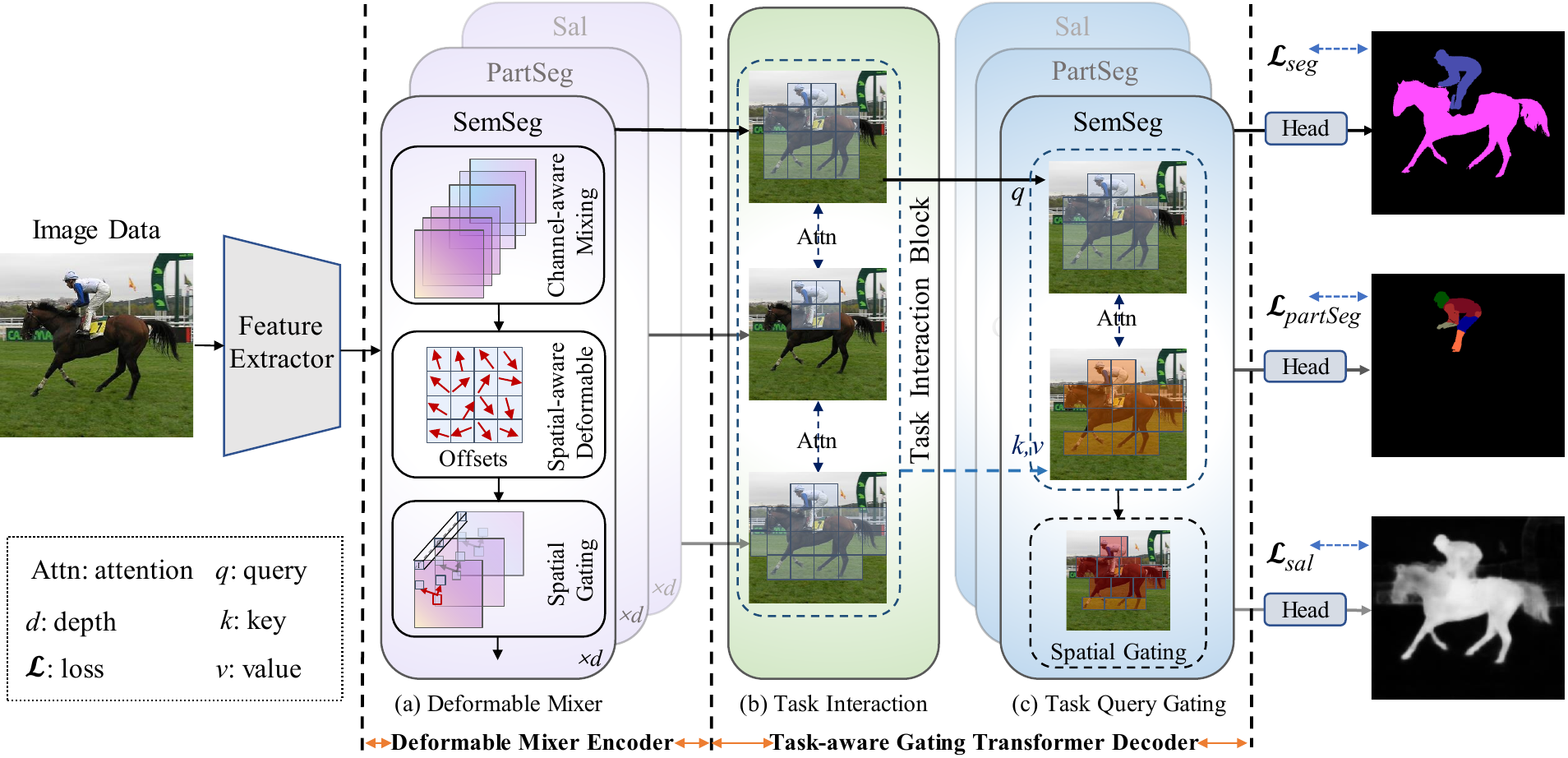}
  \caption{An overview of our model jointly handles multiple tasks with a unified encoder-decoder architecture. The DeGMix model consists of the deformable mixer encoder and task-aware gating transformer decoder. The core blocks of our DeGMix are (a) the deformable mixer (DM) that generates task-relevant deformable features, (b) the task interaction block that performs the interactions among tasks, and (c) the task query gating block that can dynamically select the task-aware features. The depth $d$ refers to the number of repetitions of the Deformable Mixer (see ablation on $d$ in Table~\ref{tab:depth}).}
  \label{fig:over_view}
\end{figure*}

\subsection{Feature Extractor}
\label{sec:fe}
The feature extractor is employed to aggregate multi-scale features and generate a shared feature map for each task.
The utilization of a shared feature map enables the model to leverage information from different scales, thereby enhancing its capacity to perform multiple tasks effectively.
As shown in Figure~\ref{fig:extractor}, the initial image data $X_{in} \in \mathbb{R}^{H \times W \times 3}$ (3 is image channel) is input to the backbone (CNN-based or Transformer-based), which then generates four stages of image features. In addition, we collect the features of four blocks when ViT is used as backbone.
The four-stage image features are up-sampled to the same resolution, and then they are concatenated along the channel dimension to obtain an image feature $X \in \mathbb{R}^{\frac{H}{4} \times \frac{W}{4} \times C}$, where $H$, $W$, and $C$ are the height, width, and channel of the image feature, respectively.
An ablation on the four-stage features is provided in Table~\ref{tab:scales}.

\subsection{Deformable Mixer Encoder}
\label{subsec:enc}

\begin{figure}[!t]
\centering
  \includegraphics[width=0.465\textwidth]{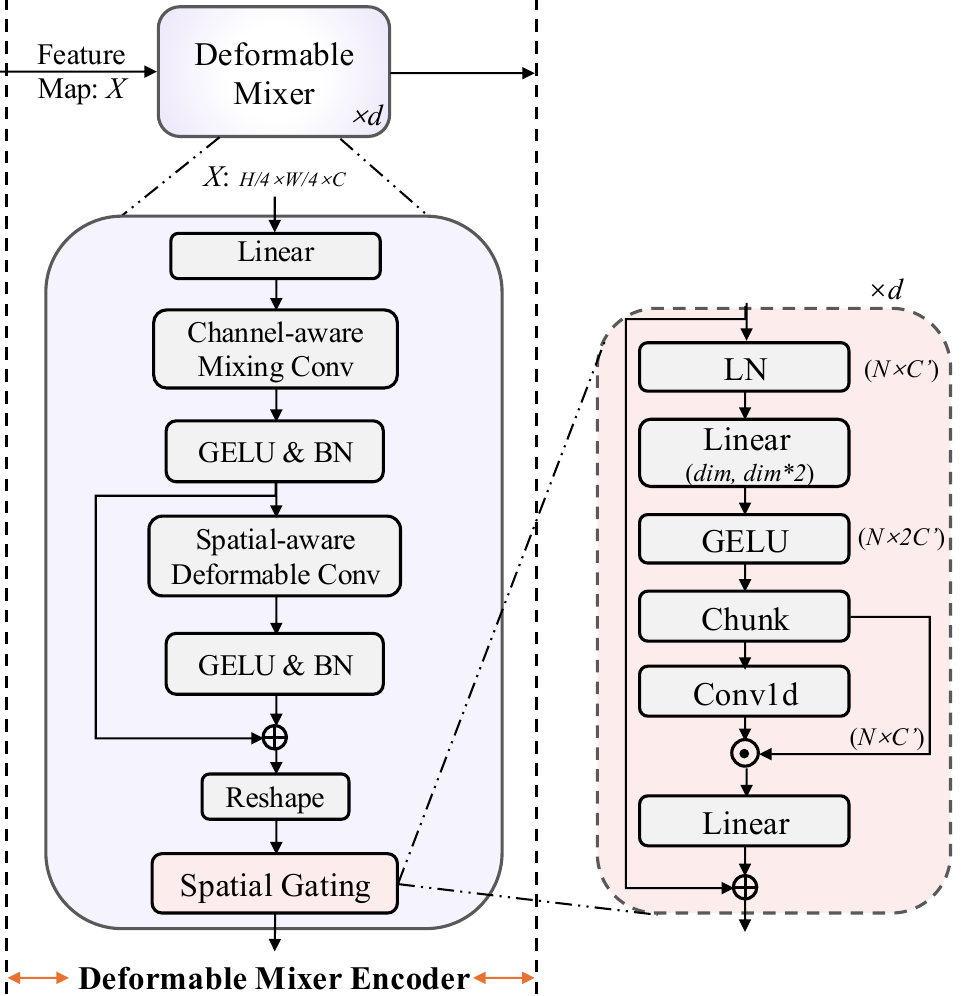}
  \caption{Illustration of our Deformable Mixer Encoder. For simplicity, we assume there is only one task ($T = 1$) in this figure.}
  \label{fig:models}
\end{figure}

\textbf{Motivation.} Inspired by Deformable ConvNets~\cite{defcnn_2019} and Deformable DETR methods~\cite{deformableDetr}, we propose the deformable mixer encoder that adaptively provides more efficient receptive fields and sampling spatial locations for each task.
To achieve this objective, we first introduce a deformable mixer encoder that consists of four operation layers, the linear, a channel-aware mixing convolution, the GELU\&BN, a spatial-aware deformable convolution, and a spatial gating.
The deformable mixer encoder separates the mixing of channel-aware location features and spatial-aware deformable features.
As shown in Figure~\ref{fig:models}, the channel-aware mixing convolution and spatial-aware deformable convolution operators are interleaved to enable interaction of both input feature dimensions (\textit{i.e.,} $(HW) \& C$). The shared spatial gating layer allows the model to learn task-agnostic features that are beneficial across all tasks, enabling better synergy between related tasks.

Specifically, the channel-aware mixing convolution learns task-specific channel dependencies, enabling adaptive information sharing across tasks, while the spatial-aware deformable convolution aggregates contextual information from a small set of learnable sampling points.
Stacking these modules sequentially yields a single deformable mixer block, whose depth effect is analyzed in Table~\ref{tab:depth}.

The deformable mixer encoder structure is shown in Figure~\ref{fig:models}.
First, a linear layer reduces the channel dimension of the image feature $X\in \mathbb{R}^{\frac{H}{4} \times \frac{W}{4} \times C}$ from $C$ to a smaller dimension $C'$.  
The linear layer can be formally expressed as:
\begin{equation}
  \begin{aligned}
    \begin{array}{ll}
        {X} = {W}\cdot \operatorname{Norm}({X}),\label{eq:linear}
    \end{array}
  \end{aligned}
\end{equation}
where Norm means LayerNorm function, after the linear layer, we obtain a smaller dimension image feature map $X\in \mathbb{R}^{\frac{H}{4} \times \frac{W}{4} \times C'}$ as the input for the downstream layers.

\noindent
\textbf{Channel-aware mixing convolution.}
Given the input image feature $X_{i,j}\in \mathbb{R}^{\frac{H}{4} \times \frac{W}{4} \times C'}$ obtained from the output of Eq.~(\ref{eq:linear}) output, the point $(i,j)$ represents the spatial location in the single channel feature map.
The channel-aware mixing convolution facilitates communication between different channels by applying the standard point-wise convolution (where the convolution kernel is 1$\times$1) to mix channel locations.
It can be formulated as follows:
\begin{equation}
X_{i, j} = W_p \cdot{X_{i, j}} + b,
\end{equation}
where the $W_p$ is the point-wise convolution weights. $b$ is a learnable bias. 
Subsequently, we add GELU activation and BatchNorm as well.
This operation is calculated as follows:
\begin{equation}\label{eq:bn01}
 X_{i, j} =  \operatorname{BN}(\sigma(X_{i, j})),
\end{equation}
where $\sigma(\cdot)$ denotes the activation function.

\noindent
\textbf{Spatial-aware deformable convolution.}
To generate the relative offsets with respect to the reference point, the image feature $X_{i,j}\in \mathbb{R}^{\frac{H}{4} \times \frac{W}{4} \times C'}$ obtained from the output of Eq.~(\ref{eq:bn01}) output, the point $(i,j)$ represents the spatial location.
$X_{i,j}$ is fed to the convolution operator to learn the corresponding offsets $\Delta_{(i,j)}$ for all reference points.
For each location point $(i,j)$ on the image feature $X_{i,j}$, the spatial deformable can be mathematically represented as follows:
\begin{equation}
D_S(X_{i,j}) = \sum_{C'=0}^{C'-1}  W_{def} \cdot {X}((i,j)+\Delta_{(i,j)}),
\end{equation}
where the $W_{def}$ is a deformable convolution weights.
The $\Delta_{(i,j)}$ is the learnable offset. 
The spatial-aware deformable convolution is followed by a GELU activation, BatchNorm, and residual connection:
\begin{equation}\label{eq:encoder}
  X_{def^\prime} = X_{i,j} + \operatorname{BN}(\sigma(D_S(X_{i,j}))),
\end{equation}
where $\sigma(\cdot)$ represents the non-linearity function (\textit{i,e.,} GELU), and BN denotes the BatchNorm operation.

We use \textit{Reshape} to denote the rearrangement function. Our feature map $X_{def'}$ can written as:
\begin{equation}\label{eq:bn1}
 X_{d'} =  \text{Reshape}(X_{def'}),
\end{equation}
where the \textit{Reshape} operation flattens the feature map $X_{d'}\in \mathbb{R}^{\frac{H}{4} \times \frac{W}{4} \times C'}$ into a sequence $\mathbb{R}^{N\times C'}$ ($N=\frac{H}{4} \times \frac{W}{4} $).
For $T$ tasks, the deformable mixer encoder generates a feature set {($X^1_{d'}, X^2_{d'}, \cdots X^T_{d'})$}, where $T$ denotes the number of tasks, as shown in Figure~\ref{fig:models}.
The task-specific features are obtained through channel-aware mixing and spatially deformable convolutions, forming the \textit{deformed features}, which are subsequently fused into the input of the shared spatial gating layer.

\noindent
\textbf{Shared spatial gating layer.}
As illustrated in Figure~\ref{fig:models}, a shared spatial gating (SSG) layer is employed to enhance information flow across tasks through a gating mechanism~\cite{d2017gatingMa}.
Gating mechanisms control the path through which information flows in the network and have proven to be useful for CNN-based and Transformer-based networks~\cite{d2017gatingMa,yang2022focalgating}. In MTL, the shared gating facilitates the exchange of information across tasks, allowing each task to leverage the information learned by the others.
Formally, given input features $x$ and $a$, a gating layer computes an output collection of feature $y$ by $y=x \cdot \sigma(a)$, where $\sigma(\cdot)$ is some activation function.
The proposed shared spatial gating is applied to refine each task independently feature through task-relevant dynamic selection.

We first apply a linear transformation to the incoming task deformed features (\textit{i.e.,} ${X_{d'}}$ from Eq.~\ref{eq:bn1}): 
\begin{align}\label{eq:ssg:encoder}
    \tilde{X}_{d'} =\ &  \sigma(\text{Linear}(\text{LN}({X_{d'}}))),
\end{align}
where LN is the layer normalization. $\sigma(\cdot)$ is the non-linearity function (\textit{i.e.,} GELU). 
After linear layer with ${X}_{d'}\in \mathbb{R}^{N \times C'}$, we obtain the feature ${\tilde{X}_{d'}} \in \mathbb{R}^{N \times 2C'}$.
The ${\tilde{X}_{d'}}$ is split into ${\tilde{X}_{d^1}}\in \mathbb{R}^{N \times C'}$ and ${\tilde{X}_{d^2}} \in \mathbb{R}^{N \times C'}$ along the channel dimension. 
Given two features $\tilde{X}_{d^1}$ and $\tilde{X}_{d^2}$, the spatial gating is formulated as follows:
\begin{align}\label{eq:ssg:cov1d:encoder}
    \tilde{X}'_{d''} =\ &  {\tilde{X}_{d^1}} \odot \text{Conv1d}(\text{LN}({\tilde{X}_{d^2}})),\\
    {X}_{d}=\ & \text{Linear}(\tilde{X}'_{d''}) +  X_{d'},
\end{align}
where LN is the layer normalization. $\odot$ is the element-wise multiplication between matrices.
\textit{Conv1d} denotes 1D convolution operation. The residual feature $X_{d'}$ comes from Eq.~\ref{eq:bn1}.
The final spatial gating feature ${X}_{d} \in \mathbb{R}^{\frac{H}{4} \times \frac{W}{4} \times C'}$ is reshaped from $\mathbb{R}^{N \times C'}$ ($N=\frac{H}{4} \times \frac{W}{4} $) via \textit{Reshape} operation.
After the shared spatial gating layer, we collect the $T$ tasks' deformed feature set {(${X}^1_{d}, {X}^2_{d}, \cdots {X}^T_{d})$}.

The SSG improves the exchange of information across tasks, enabling each to benefit from others’ representations.
Its task-specific outputs are refined by a deformable mixer that generates \textit{deformed features}, which are then integrated into the downstream task-interaction and task-query blocks.

\subsection{Task-aware Gating Transformer Decoder}
\label{subsec:dec}
\begin{figure}[!t]
\centering
  \includegraphics[width=0.47\textwidth]{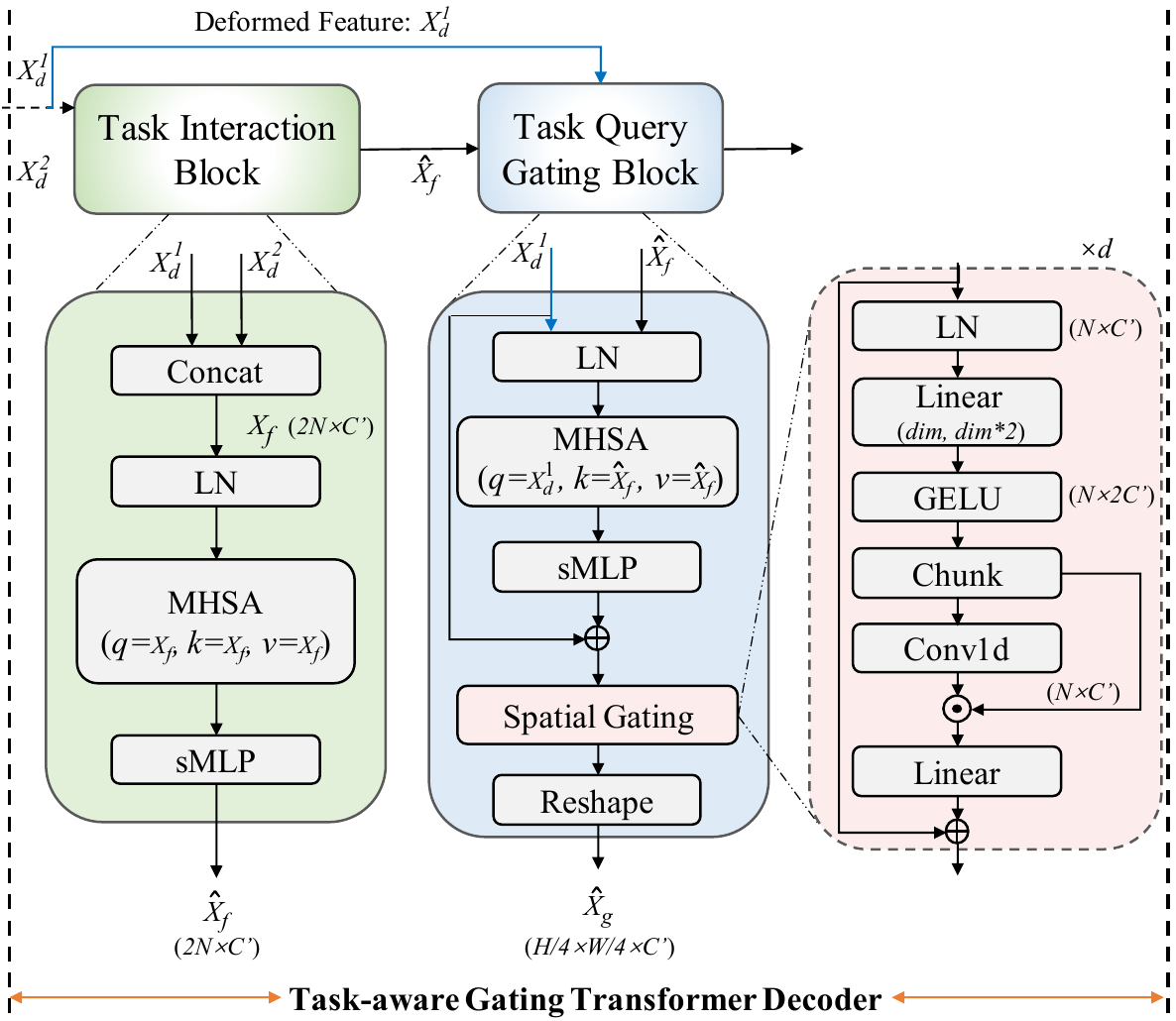}
  \caption{Illustration of our Task-aware Gating Transformer Decoder. The $dim$ denotes the input dimension. \textit{Conv1d} denotes 1D convolution operation. For simplicity, we assume there is two tasks in the task interaction block and one task in the task query gating block.}
  \label{fig:decoder_gating}
\end{figure}

In the task-aware gating transformer decoder, we design the task interaction block and task query gating block (see Figure~\ref{fig:decoder_gating}). 
To effectively model inter-task dependencies, the task interaction block employs a self-attention mechanism composed of a multi-head self-attention (MHSA) module and a lightweight multi-layer perceptron (sMLP). Similarly, the downstream task query block also integrates MHSA and sMLP modules but differs in the nature of its query features, which are tailored to task-specific representations. 
The feature is projected into the queries (\textit{Q}), keys (\textit{K}), and values (\textit{V}) of dimension $d_k$, and self-attention is being computed by the \textit{Q}, \textit{K} and \textit{V}.
The self-attention operator is calculated as follows:
\begin{align}\label{eq:mhsa}
    &\text{MHSA}(Q, K, V) = \text{softmax}(\frac{QK^\mathcal{T}}{\sqrt{d_k}})V,
\end{align}
where $Q \in \mathbb{R}^{N \times C'}$, $K \in  \mathbb{R}^{N \times C'}$ and $ V \in  \mathbb{R}^{N \times C'}$ are the query, key and value matrices; $\text{MHSA}(Q, K, V) \in \mathbb{R}^{N \times C'}$.

\noindent
\textbf{Task interaction block.}
As shown in Figure~\ref{fig:models} (green box), we first concatenate the deformed features from all output of the deformable mixer encoder.

\begin{equation}
 X_f =  \operatorname{Concat}(X^1_d, X^2_d, \cdots X^T_d),
\end{equation}
where $X_f \in \mathbb{R}^{TN \times C'}$ is the fused feature. The $T$ means task number in $X^T_d \in \mathbb{R}^{N \times C'}$.
Then, for efficient task interaction, we construct a self-attention strategy via the fused feature $X_f$:
\begin{align}\label{eq:mhsafusion}
    X_f' =\ &\text{MHSA}(Q,K,V=\operatorname{LN}(X_f),\operatorname{LN}(X_f),\operatorname{LN}(X_f)),\\
    \hat X_f =\ &\operatorname{sMLP}(X_f'),
\end{align}
where $\hat X_f \in \mathbb{R}^{TN \times C'}$ is the task-interacted feature. LN means LayerNorm. sMLP consists of a linear layer and a LayerNorm.

\noindent
\textbf{Task query gating block.}
As illustrated in Figure~\ref{fig:decoder_gating} (left), we take the deformed feature $X_d$ as task query and the task-interacted feature $\hat X_f$ as key \& value to MHSA.
The deformed feature is applied as a query in MHSA to decode the task awareness feature from the task-interacted feature for each task prediction.
We first apply the LayerNorm in parallel to generate queries $Q$, keys $K$, and values $V$:

\begin{align}\label{eq:taskqueryln}
   \hat Q=\operatorname{LN}(X_d),\quad
   \hat K=\operatorname{LN}(\hat X_f),\quad
   \hat V=\operatorname{LN}(\hat X_f),
\end{align}
where LN is the layer normalization. $X_d$ and $\hat X_f$ are the output of the deformable mixer encoder and the task interaction block, respectively. 
Then, the task query block operation using a MHSA is calculated as follows:
\begin{align}\label{eq:taskquery}
    \hat X_d = \text{MHSA}(\hat Q, \hat K, \hat V),\\
    \hat{X} = X_d + \operatorname{sMLP}(\hat X_d),
\end{align}
where the residual feature $X_d$ comes from Eq.~(9).
The task awareness feature $\hat{X} \in \mathbb{R}^{N \times C'}$.

\noindent
\textbf{Shared spatial gating layer.}
As illustrated in Figure~\ref{fig:decoder_gating}, we use a shared spatial gating (SSG) layer with a gating mechanism~\cite{d2017gatingMa} in decoder to allow the tasks to interact and share important features from each other. 
The SSG process can be formulated as follows:
\begin{align}\label{eq:ssg:ln}
    \tilde{X}_g =\ &  \sigma(\text{Linear}(\text{LN}(\hat{X}))),
\end{align}
where LN is the layer normalization. $\sigma(\cdot)$ is the non-linearity function (\textit{i.e.,} GELU). 
We first apply a linear transformation to the incoming task awareness feature (\textit{i.e.,} $\hat{X}$).
After Linear with $\hat{X}\in \mathbb{R}^{N \times C'}$, we obtain the feature ${\tilde{X}_g} \in \mathbb{R}^{N \times 2C'}$.
The ${\tilde{X}_g}$ is split into ${X_{g^1}}\in \mathbb{R}^{N \times C'}$ and ${X_{g^2}} \in \mathbb{R}^{N \times C'}$ along the channel dimension. 
Given two features $X_{g^1}$ and $X_{g^2}$, the spatial gating is formulated as:
\begin{align}\label{eq:ssg:cov1d}
    X'_g =\ &  {X_{g^1}} \odot \text{Conv1d}(\text{LN}({X_{g^2}})),\\
    \hat{X}_g=\ & \text{Reshape}(\text{Linear}(X'_g) + \hat{X}),
\end{align}
where $\odot$ is the element-wise multiplication between matrices.   
\textit{Conv1d} denotes 1D convolution operation.
The residual feature $\hat{X}$ comes from Eq.~(16).
The final spatial gating feature $\hat{X}_g \in \mathbb{R}^{\frac{H}{4} \times \frac{W}{4} \times C'}$ is reshaped from $\mathbb{R}^{N \times C'}$ ($N=\frac{H}{4} \times \frac{W}{4} $) via \textit{Reshape} operation.
Overall, the SSG controls the task information flow by gating through the task awareness features, thereby allowing each task to focus on the fine details complementary to the other tasks.
By leveraging this gating mechanism, the SSG can compute nonlinear interactions between two input features, which significantly enriches the expressiveness of the dynamically selected features.
To further demonstrate the advantages of the shared spatial gating layer, we conduct an ablation study, as shown in Table~\ref{tab:module}.

DeGMix integrates a deformable mixer encoder and a task-aware gating Transformer decoder.
The encoder adaptively focuses on task-relevant spatial regions, enhancing localization-sensitive tasks such as boundary detection. The decoder promotes cross-task interaction while preserving task specificity, aided by a shared spatial gating layer that maintains coherence and efficiency across tasks.

\subsection{MTL Loss Function}
For balancing the loss contribution for each task, we set the weight $\alpha_t$ to decide the loss contribution for the task $t$. A weighted sum $\mathcal L_{total}$ of task-specific losses:
	\begin{equation}\label{equ:loss}
	\begin{split}
	    \begin{aligned}
    {\mathcal L_{total}} &= \sum_{t=1}^{T} \alpha_{t}{\mathcal L_{t}}, \\
       \end{aligned}
    \end{split}
	\end{equation}
where the $\mathcal L_{t}$ is a loss function for task $t$.
For fair comparison, both $\alpha_t$ and $\mathcal{L}_t$ follow the settings in ATRC~\cite{atrc_2021} and MQTransformer~\cite{mqtrans_xy}.
Let $\mathcal L_{total}$ be the set of total loss of the MTL framework ($\mathcal L_{total} = \{\mathcal L_{seg}, \mathcal L_{partseg}, \mathcal L_{sal}, \mathcal L_{depth}, \mathcal L_{normal}, {\mathcal L_{bound}}\}$).
Cross entropy loss is used for SemSeg ($\mathcal L_{seg}$), Human Parts Seg ($\mathcal L_{partseg}$) and saliency ($\mathcal L_{sal}$) tasks, which computes the cross entropy loss between input and target.
L1 loss is adopted for Depth ($\mathcal L_{depth}$) and Surface Normal ($\mathcal L_{normal}$) tasks.
Binary cross entropy loss is employed for Boundary (${\mathcal L_{bound}}$), which measures binary cross entropy between target and input logits.

\section{Experiments}
\label{sec:exp}

\subsection{Setup}
\noindent
\textbf{Implementation.}
All Swin-Transformer backbones generate four feature scales (1/4, 1/8, 1/16, 1/32) that are aggregated through a simple multi-scale feature extractor (see Section \ref{sec:fe}). For ViT-B/L, we extract features from every three and six transformer blocks, respectively (\textit{i.e.,} the 3rd–6th–9th–12th layers for ViT-B and the 6th–12th–18th–24th for ViT-L). DeGMix is trained using SGD with a learning rate of $10^{-3}$ and weight decay of $5\times10^{-4}$.
Unless specified otherwise, all models are trained for 400 epochs.
The task weights in Eq. (\ref{equ:loss}) are set as $\alpha_{seg}$ = 1.0, $\alpha_{depth}$ = 1.0, $\alpha_{partseg}$ = 2.0, $\alpha_{sal}$ = 5.0, $\alpha_{normal}$ = 10.0, and $\alpha_{bound}$ = 50.0 to balance multi-task objectives. All experiments are conducted on 8 A100 GPUs.

\noindent
\textbf{Datasets.}
We conduct experiments on three publicly accessible datasets, NYUD-v2~\cite{NYUD2012}, PASCAL-Context~\cite{pascal2014}, and Cityscapes~\cite{2016cityscapes}. 
NYUD-v2 contains paired RGB–Depth images with 795 training and 654 testing samples, and is widely used for semantic segmentation (SemSeg), depth estimation (Depth), surface normal estimation (Normal), and boundary detection (Bound).
PASCAL-Context contains 10103 training/validation and 9637 testing images, offering dense annotations for the entire scene across tasks such as SemSeg, human-part segmentation (PartSeg), saliency estimation (Sal), Normal, and Bound. Cityscapes focuses on urban street understanding and is commonly adopted for SemSeg and depth estimation.

\noindent
\textbf{Metrics.}
We adopt six standard metrics to evaluate our model against prior multi-task approaches: mean Intersection over Union (mIoU), root mean square error (rmse), mean Error (mErr), absolute error (aErr), optimal dataset scale F-measure (odsF), and maximum F-measure (maxF).
These metrics provide comprehensive assessments of our model's performance across various tasks and datasets.
Since, for multi-task learning, it’s difficult to measure the performance of a model with a single metric; we introduce the average per-task performance drop ($\Delta_m$) metric.
$\Delta_m$ is used to quantify multi-task performance. $\Delta_m=\frac{1}{T} \sum_{i=1}^T(F_{m,i}-F_{s,i})/F_{s,i}\times100\%$, where $m$, $s$ and $T$ mean multi-task model, single task baseline and task numbers.
$\Delta_m$: the higher, the better.

\begin{table*}[!tp]
\small
\caption{Comparison of the MTL models with state-of-the-art on NYUD-v2 dataset. The notation '$\downarrow$': lower is better. The notation '$\uparrow$': Higher is better. $\Delta_m$ denotes the average per-task performance drop. "Params" denotes parameters. Swin-$\diamond$ indicates that the specific Swin model is uncertain. $\dagger$ denotes results not reported in InvPT~\cite{InvPT_2022} but from our evaluation using the official code.}\label{tab:stoa_nyud_v2}
\centering
\begin{tabular}{llllllllr}
\toprule[0.1em]
 \multirow{2}*{Model}  & \multirow{2}*{Backbone}  &Params  &FLOPs  & {SemSeg}   &Depth & Normal & Bound &\multirow{2}*{$\Delta_m[\%]$$\uparrow$}\\
      &  &(M) &(G) &(mIoU)$\uparrow$   &(rmse)$\downarrow$   & (mErr)$\downarrow$  &(odsF)$\uparrow$\\
\hline
single task baseline &HRNet18   &16.09 &40.93  &38.02 &0.6104 &20.94 &76.22 &0.00\\
multi-task baseline  &HRNet18   &4.52  &17.59  &36.35 &0.6284 &21.02 &76.36 &-1.89\\
Cross-Stitch\cite{Cross-Stitch_2016} &HRNet18 &4.52 &17.59 &36.34  &0.6290  &20.88  &76.38 &-1.75\\
Pad-Net\cite{Pad-net_2018}           &HRNet18 &5.02 &25.18 &36.70  &0.6264  &20.85  &76.50 &-1.33\\
PAP\cite{PAP-Net_2019}               &HRNet18 &4.54 &53.04 &36.72  &0.6178  &20.82  &76.42 &-0.95\\
PSD\cite{pattern_struct_diffusion_2020} &HRNet18 &4.71 &21.10 &36.69 &0.6246 &20.87 &76.42 &-1.30\\
NDDR-CNN\cite{NDDR-CNN_2019}          &HRNet18  &4.59  &18.68 &36.72 &0.6288 &20.89 &76.32 &-1.51\\
MTI-Net\cite{Mti-net_2020}            &HRNet18  &12.56 &19.14 &36.61 &0.6270 &20.85 &76.38 &-1.44\\
ATRC\cite{atrc_2021}                  &HRNet18  &5.06  &25.76 &38.90 &0.6010 &20.48 &76.34 &1.56\\
DeMT                                  &HRNet18  &4.76  &22.07 &39.18 &0.5922 &20.21 &76.40 &2.37\\
\rowcolor{mygray}DeGMix (Ours)        &HRNet18  &4.92  &28.83 &39.32 &0.5927 &20.17 &76.40 &2.50\\

\hdashline
single task baseline            &Swin-T  &115.08 &161.25 &42.92   &0.6104  &20.94  &76.22 &0.00\\
multi-task baseline             &Swin-T  &32.50  &96.29  &38.78   &0.6312  &21.05  &75.60 &-3.74\\
MQTransformer\cite{mqtrans_xy}  &Swin-T  &35.35  &106.02 &43.61   &0.5979  &20.05  &76.20 &0.31\\
InvPT\cite{InvPT_2022}          &Swin-T  &60.14$^{\dagger}$  &162.52$^{\dagger}$ &44.27   &0.5589  &20.46  &76.10 &3.35$^{\dagger}$\\
InvPT++\cite{invpt++}           &Swin-T  &59.95$^{\dagger}$  &172.29$^{\dagger}$ &44.94   &0.5554  &20.43  &76.20 &3.75$^{\dagger}$ \\
DeMT                            &Swin-T  &32.07  &100.70 &46.36   &0.5871  &20.65  &76.90 &3.36\\
\rowcolor{mygray}DeGMix (Ours)   &Swin-T  &33.50  &125.49  &47.30  &0.5630  &20.14  &77.20 &5.10\\

\hdashline
single task baseline            &Swin-S  &200.33 &242.63 &48.92  &0.5804  &20.94  &77.20  &0.00\\
multi-task baseline             &Swin-S  &53.82  &116.63 &47.90  &0.6053  &21.17  &76.90  &-1.96\\
MQTransformer\cite{mqtrans_xy}  &Swin-S  &56.67  &126.37 &49.18  &0.5785  &20.81  &77.00  &1.59\\
MTFormer\cite{xu2022mtformer}   &Swin-$\diamond$&64.03&117.73&50.56 &0.4830 &-&-          &4.12\\
DeMT                            &Swin-S  &53.03  &121.05  &51.50  &0.5474  &20.02  &78.10 &4.12\\
\rowcolor{mygray}DeGMix (2-task) (Ours)&Swin-S &52.16  &96.73 &53.59 &0.5521 &-     &-     &6.5\\
\rowcolor{mygray}DeGMix (Ours)   &Swin-S  &54.82  &145.84 &52.36	&0.5430	&19.99	&78.50  &4.83\\
\hdashline
single task baseline         &Swin-L  &789.96  &819.93 &56.46	&0.508	&19.38	&78.80 &0.00 \\
multi-task baseline          &Swin-L  &204.96  &316.87 &54.53	&0.532	&19.51	&78.30 &-2.36 \\
InvPT\cite{InvPT_2022}       &Swin-L  &292.7$^{\dagger}$ &417.27$^{\dagger}$ &51.76 &0.5020 &19.39  &77.60 &-2.22$^{\dagger}$\\
InvPT++\cite{invpt++} &Swin-L &292.1$^{\dagger}$ &451.3$^{\dagger}$ &52.35 &0.4921 &18.99 &77.90 &-0.82$^{\dagger}$ \\
\rowcolor{mygray}DeGMix (Ours) &Swin-L  &202.92 &321.22 &57.55	&0.5037	&19.21	&79.00  &0.98\\
\hdashline
InvPT\cite{InvPT_2022} &ViT-B  &160.82$^{\dagger}$ &327.61$^{\dagger}$ &50.30 &0.5367 &19.00 &77.60 &-\\
InvPT++\cite{invpt++}  &ViT-B &159.94$^{\dagger}$ &359.93$^{\dagger}$ &49.83 &0.5303 &18.83 &77.20 &-\\
\rowcolor{mygray}DeGMix (Ours) &ViT-B &114.33  &321.04 &52.09 &0.5467 &20.34 &77.90 &-\\
InvPT\cite{InvPT_2022} &ViT-L &402.1$^{\dagger}$ &555.57$^{\dagger}$ &53.56 &0.5183 &19.04 &78.10 &-\\
InvPT++\cite{invpt++}  &ViT-L &401.3$^{\dagger}$ &589.49$^{\dagger}$   &53.89 &0.5091 &18.65 &78.10  &-\\
TaskPrompter\cite{taskprompter2023}&ViT-L &513.1$^{\dagger}$ &789.90$^{\dagger}$    &55.30 &0.5152 &18.47 &78.20  &-\\
\rowcolor{mygray}DeGMix (Ours)      &ViT-L &334.1  &507.72  &56.29 &0.5067 &19.34 &78.90 &-\\
\bottomrule[0.1em]
  \end{tabular}
\end{table*}

\noindent
\textbf{Backbones.}
We evaluate our method using several CNN and Vision Transformer backbones: HRNetV2-W18-small (HRNet18), HRNetV2-W48 (HRNet48)~\cite{HRnet_19}, Swin-Tiny (Swin-T), Swin-Small (Swin-S), Swin-Base (Swin-B), Swin-Large (Swin-L)~\cite{swin}, ViT-Base (ViT-B), and ViT-Large (ViT-L)~\cite{ViT2021}.

\noindent
\textbf{Baselines.}
We categorize baselines into two groups: multi-task learning (MTL) and single-task learning (STL). These serve as reference points to evaluate the proposed DeGMix framework.
DeGMix employs both CNN-based (\textit{e.g.}, HRNet) and Transformer-based (\textit{e.g.}, Swin-Transformer, ViT) backbones, including HRNet18, Swin-T/S/B/L, and ViT-B/L, for experiments on NYUD-v2 and PASCAL-Context.
We also compare with representative Transformer-based MTL approaches, such as InvPT~\cite{InvPT_2022}, MTFormer~\cite{xu2022mtformer}, and InvPT++~\cite{invpt++}.
For fine-tuning settings, “single-task full fine-tuning” denotes training a separate pre-trained model per task, while “MTL-tuning decoders only” fixes the encoder and updates only decoder parameters.
DeGMix follows the latter strategy.
We further include parameter-efficient tuning baselines: Adapter~\cite{he2022adapter}, which adds task-specific modules to each Transformer layer; BitFit~\cite{bitfit2022}, which tunes only bias and patch layers; and LoRA~\cite{hu2022lora}, which applies low-rank adaptation (rank=4, scale=4) to attention layers.

\subsection{Comparison with the State-of-the-art}

\begin{table*}[!ht]
\small
\centering
\caption{Comparison of the MTL models with state-of-the-art on PASCAL-Context dataset. The notation ‘$\downarrow$’: lower is better. The notation ‘$\uparrow$’: Higher is better. $\Delta_m$ denotes the average per-task performance drop (the higher, the better). Swin-$\diamond$ indicates that the specific Swin model is uncertain. $\dagger$ means we test.}
\label{tab:stoa_pascal}
\begin{tabular}{lllllllr}
\toprule[0.1em]
 \multirow{2}*{Model}  & \multirow{2}*{Backbone}  & {SemSeg}   &PartSeg &Sal  & Normal & Bound &\multirow{2}*{$\Delta_m[\%]$$\uparrow$}\\
     &  & (mIoU)$\uparrow$  & (mIoU)$\uparrow$  &(maxF)$\uparrow$  &(mErr)$\downarrow$  &(odsF)$\uparrow$\\
\hline
single task baseline  &HRNet18   &62.23  &61.66 &85.08  &13.69 &73.06 &0.00\\
multi-task baseline   &HRNet18   &51.48  &57.23 &83.43  &14.10 &69.76 &-6.77\\

PAD-Net~\cite{Pad-net_2018}       &HRNet18  &53.60 &59.60 &65.80 &15.30 &72.50 &-4.41\\
ATRC~\cite{atrc_2021}             &HRNet18  &57.89 &57.33 &83.77 &13.99 &69.74 &-4.45\\
MQTransformer\cite{mqtrans_xy}    &HRNet18  &58.91 &57.43 &83.78 &14.17 &69.80 &-4.20\\
DeMT                              &HRNet18  &59.23	&57.93 &83.93 &14.02 &69.80 &-3.79\\
\rowcolor{mygray}DeGMix (Ours)    &HRNet18  &59.56	&57.71 &84.03 &14.00 &69.70 &-3.40\\
\hdashline
single task baseline               &Swin-T  &67.81	&56.32  &82.18  &14.81  &70.90 &0.00\\
multi-task baseline                &Swin-T   &64.74	 &53.25  &76.88  &15.86  &69.00 &-3.23\\
MQTransformer\cite{mqtrans_xy}     &Swin-T   &68.24	 &57.05	 &83.40  &14.56  &71.10 &1.07\\
DeMT                               &Swin-T   &69.71	 &57.18  &82.63  &14.56  &71.20 &1.75\\
\rowcolor{mygray}DeGMix (Ours)     &Swin-T   &69.44	 &58.02	 &83.31	 &14.31  &71.20 &1.94\\
\hdashline
single task baseline               &Swin-S   &70.83  &59.71  &82.64  &15.13  &71.20 &0.00\\
multi-task baseline                &Swin-S   &68.10  &56.20  &80.64  &16.09  &70.20 &-3.97\\
MQTransformer\cite{mqtrans_xy}     &Swin-S   &71.25  &60.11  &84.05  &14.74  &71.80 &1.27\\
DeMT                               &Swin-S   &72.01  &58.96  &83.20  &14.57  &72.10 &1.36\\
\rowcolor{mygray}DeGMix (Ours)     &Swin-S   &71.54  &61.49	 &83.70	 &14.90  &72.20 &1.62\\
\hdashline
single task baseline               &Swin-B   &74.91  &62.13 &82.35  &14.83  &73.30  &0.00 \\
multi-task baseline                &Swin-B   &73.83  &60.59 &80.75  &16.35  &71.10  &-3.81\\
MTFormer\cite{xu2022mtformer} &Swin-$\diamond$ &74.15 &64.89 &67.71  &-&-      &2.41\\
DeMT                               &Swin-B    &75.33  &63.11 &83.42  &14.54  &73.20  &1.04\\
\rowcolor{mygray}DeGMix (3-task) (Ours)&Swin-B &76.77 &66.05 &83.53  &-  &-  &3.0\\
\rowcolor{mygray}DeGMix (Ours)       &Swin-B   &75.37 &64.82	&83.75	&14.22	&73.0   &1.84\\
\hdashline
single task baseline              &Swin-L  &79.26	&68.92  &83.84  &14.28  &74.50  &0.00\\
multi-task baseline               &Swin-L  &77.35	&63.86  &82.87  &14.84  &73.10  &-3.33\\
{InvPT}\cite{InvPT_2022}          &Swin-L  &78.53   &68.58  &83.71  &14.56  &73.60  &-0.947$^\dagger$ \\
\rowcolor{mygray}DeGMix (Ours)     &Swin-L  &78.54	&67.42	&83.74	&14.17	&74.90 	&-0.38 \\
\bottomrule[0.1em]
  \end{tabular}
\end{table*}

\noindent
\textbf{NYUD-v2.}
We evaluate DeGMix on the NYUD-v2~\cite{NYUD2012} dataset, a widely used benchmark for multi-task dense prediction.
Following the standard evaluation protocol~\cite{NYUD2012,mqtrans_xy}, we compare with state-of-the-art methods across multiple backbones, including HRNet18, Swin-T/S/B/L, and ViT-B/L.
As shown in Table~\ref{tab:stoa_nyud_v2}, DeGMix consistently surpasses previous CNN- and Transformer-based models such as MQTransformer~\cite{mqtrans_xy}, MTFormer~\cite{xu2022mtformer}, InvPT~\cite{InvPT_2022}, and InvPT++~\cite{invpt++}.
In addition, we also observe that using a Transformer as a backbone model is more promising compared to CNN as the backbone.
Because Transformer-based and CNN-based models use similar Parameters, the former shows higher accuracy in all metrics.
Specifically, our DeGMix using Swin-T obtains 47.30 mIoU (SemSeg) accuracy, which is +3.69 (47.30 $vs.$ 43.61) higher than that of MQTransformer with the same Swin-T backbone and similar FLOPs, while being more parameter efficient.
We also find that DeGMix can substantially outperform InvPT~\cite{InvPT_2022} by 3.03 mIoU SemSeg (47.30 $vs.$ 44.27), and we note DeGMix still beats it by +1.75 on $\Delta_m$ metric.
MuIT~\cite{transfmult2022} shows an improvement with a 13.3\% increase in relative performance for semantic segmentation and an 8.54\% increase for depth tasks.
While our origin DeMT achieves a 14.74\% and 9.43\% increase compared to MuIT~\cite{transfmult2022}.
As shown in Figure~\ref{fig:params}, our method can outperform the state-of-the-art InvPT, which requires 4.3$\times$ more parameters compared to DeGMix.
As expected, our DeGMix using Swin-L performs the best, with an almost +5.79 (57.55 $vs.$ 51.76) mIoU improvement in the SemSeg task compared to InvPT~\cite{InvPT_2022}.
In addition, we also report the performance of InvPT\cite{InvPT_2022} and InvPT++\cite{invpt++} with ViT-B and ViT-L~\cite{ViT2021} backbones, respectively.
These results demonstrate that DeGMix achieves consistent gains across diverse multi-task benchmarks, surpassing prior methods with fewer parameters.
The model exhibits strong flexibility and generalization, setting new state-of-the-art results on NYUD-v2~\cite{NYUD2012} and outperforming both CNN- and Transformer-based counterparts across all metrics.

\begin{table}[!t]
  \caption{ Comparison with MTL models on Cityscapes. We report the results for semantic segmentation (SemSeg) and depth estimation (Depth) tasks. $\Delta_m$ denotes the average per-task performance drop (higher is better). ‘$\downarrow$’: Lower is better. Notation ‘$\uparrow$’: Higher is better. The best numbers are in \textbf{bold}.}
  \label{tab:cityscapes}
  \centering
\begin{tabular}{llllllllll}
\toprule[0.1em]
 \multirow{2}*{Model}     & {SemSeg}   &Depth &$\Delta_m[\%]$  \\
      & (mIoU)$\uparrow$   &(aErr)$\downarrow$    &(\%)$\uparrow$\\
\noalign{\smallskip}
\hline
\noalign{\smallskip}
Sing-task Learning (STL)                                    &75.92     &0.0119  &0.00\\
\hdashline
Uncertainty~\cite{2018MTL_uncertainty} &74.95     &0.0121  &-1.73\\
GradNorm~\cite{chen2018gradnorm}       &74.88     &0.0123  &-2.70\\
DWA~\cite{liu2019MTL_DWA}              &75.39     &0.0121  &-1.42\\
PCGrad~\cite{yu2020_MTL_pcgrad}        &75.62     &0.0122  &-1.73\\
GradDrop~\cite{chen2020_MTL_graddrop}  &75.69     &0.0123  &-2.15\\
IMTL-H~\cite{l2021_IMTL}               &75.33     &0.0120  &-1.20\\
CAGrad~\cite{liu2021_MTL_CAgrad}       &75.45     &0.0124  &-2.69\\
\hdashline
DeGMix (Ours)                           &\textbf{77.10} &\textbf{0.0118} &\textbf{1.18}\\
\bottomrule[0.1em]
\end{tabular}
\end{table}

\noindent
\textbf{PASCAL-Context.}
We evaluate DeGMix on the PASCAL-Context~\cite{pascal2014} dataset, a widely used benchmark for dense prediction. Table~\ref{tab:stoa_pascal} reports results across five backbones (HRNet18, Swin-T/S/B/L).
Our model obtains significantly better results when compared with the baseline and other models. 
For example, DeGMix improves MQTransformer~\cite{mqtrans_xy} with the same Swin-T backbone by +1.2 mIoU in SemSeg.
Our DeGMix using Swin-T achieves the best performance among models on several metrics and can reach an average per-task performance drop ($\Delta_m$) performance of 1.94 \%.
Compared to the MTFormer~\cite{xu2022mtformer}, our DeGMix (3-task) significantly outperforms it on the same three tasks (\textit{i.e.,} Semseg, PartSeg, and Sal).
In addition, when using HRNet18 and Swin-L as the backbone, our DeGMix also achieves the best performance on the $\Delta_m$ metric.
We visualize qualitative results on PASCAL-Context in Figure \ref{fig:vis_diff_module}.
DeGMix consistently outperforms InvPT across datasets, confirming its effectiveness and generalization under both CNN- and Transformer-based backbones.

\begin{table*}[!t]
\caption{We report the evaluation results for our DeGMix on NYUD-v2 dataset. Three sets are '$S-D$' (segmentation + depth), '$S-D-N$' (+ normals), and '$S-D-N-B$' (+ boundary). Notation '$\downarrow$': lower is better. Notation '$\uparrow$': Higher is better. \label{tab:diff_task_test_nyud_v2}}
\centering
\begin{tabular}{lllclllllll}
\toprule[0.1em]
\multirow{2}*{Exp.} &\multirow{2}*{Model} &\multirow{2}*{Trained} &\centering{Tested}  &SemSeg &Depth &Normal &Bound \\
       & & &(+ (Task) is on new task) &(mIoU)$\uparrow$ &(rmse)$\downarrow$  &(mErr)$\downarrow$  &(odsF)$\uparrow$\\
\hline
1 &\multicolumn{8}{c}{Multi-task learning is trained 400 epochs }\\
\hline
1.1 &DeGMix   &'$S-D$'     &\makebox[20ex][l]{'$S-D$'}   &47.98   &0.5707 &-      &- \\
1.2 &DeGMix   &'$S-D-N$'   &\makebox[20ex][l]{'$S-D-N$'} &47.07   &0.5640 &20.00  &-   \\
1.3 &DeGMix   &'$S-D-B$'   &\makebox[20ex][l]{'$S-D-B$'} &47.60   &0.5737 &-      &77.6 \\
1.4 &DeGMix   &'$S-N-B$'   &\makebox[20ex][l]{'$S-N-B$'} &47.00   &-      &20.18  &77.2 \\
1.5 &DeGMix   &'$D-N-B$'   &\makebox[20ex][l]{'$D-N-B$'} &-       &0.5732 &20.04  &77.0 \\
1.6 &DeGMix   &'$S-D-N-B$' &\makebox[20ex][l]{'$S-D-N-B$'} &47.30 &0.5630 &20.14  &77.2 \\
\hline
2 &\multicolumn{8}{c}{Multi-task learning, adding of task fine-tuning with 100 epochs}\\
\hline
2.1 &DeGMix  &'$S-D$'     &\makebox[20ex][l]{'$S-D \ (+N)$'}  &47.59 &0.5772  &\orange{21.06}  &- \\
2.2 &DeGMix  &'$S-D$'     &\makebox[20ex][l]{'$S-D \ (+N, B)$'} &47.47 &0.5738  &\orange{21.01}  &\orange{76.7}  \\
2.3 &DeGMix  &'$S-D-N$'   &\makebox[20ex][l]{'$S-D-N \ (+B)$'}  &46.63 &0.5653  &19.99  &\orange{76.6}  \\
2.4 &DeGMix  &'$S-D-N-B$' &\makebox[20ex][l]{'$S-D-N-B$'}       &47.30  &0.5630 &20.14  &77.2   \\
\hline
3 &\multicolumn{8}{c}{Multi-task learning, leave one task out format, addition of the left out task}\\
\hline
3.1 &DeGMix  &'$S-D-N$'   &\makebox[20ex][l]{'$S-D-N\ (+B)$'}   &46.72  &0.5671 &20.12          &\orange{29.3}  \\
3.2 &DeGMix  &'$S-D-B$'   &\makebox[20ex][l]{'$S-D-B\ (+N)$'}   &47.03  &0.5717 &\orange{100.66} &22.5 \\
3.3 &DeGMix  &'$S-N-B$'   &\makebox[20ex][l]{'$S-N-B\ (+D)$'}   &46.72  &\orange{3.0126} &47.42  &21.2  \\
3.4 &DeGMix  &'$D-N-B$'   &\makebox[20ex][l]{'$D-N-B\ (+S)$'}   &\orange{0.475} &2.2626	&50.26  &14.0  \\
\bottomrule[0.1em]
  \end{tabular}
\end{table*}

\noindent
\textbf{Results on Cityscapes.}
To further demonstrate the robustness of our method, we also evaluate DeGMix on the Cityscapes dataset. We present the results in Table~\ref{tab:cityscapes}. Specifically, for the Cityscapes 2-task scenario, we report the performance for semantic segmentation (SemSeg), depth estimation (Depth), and the average per-task performance drop (($\Delta_m$)). Our experiments consistently show that DeGMix outperforms existing approaches across all metrics. Notably, our method achieves a significant improvement of +1.18 ($\Delta_m$) over the STL baseline, a trend that mirrors the performance gains observed on the NYUD-v2 and PASCAL-Context datasets. These results highlight the effectiveness and generalizability of DeGMix across diverse dense prediction tasks.

\subsection{MTL Performs Comparable to Unseen Tasks}
To evaluate the generalization of our MTL framework, we test it on unseen tasks from the NYUD-v2 dataset (Table~\ref{tab:diff_task_test_nyud_v2}). We design two experimental settings: augmenting tasks (Exp. 1–2) and leave-one-out (Exp. 3). All Exp. 2 models are fine-tuned for 100 epochs, while Exp. 3 uses a leave-one-task-out model to assess performance across all four tasks.

\noindent
\textbf{The effects of unseen task fine-tuning.}
Table~\ref{tab:diff_task_test_nyud_v2} presents the results of fine-tuning DeGMix on unseen tasks, where “$+${Task}” denotes training on existing tasks followed by adaptation to a new one. Across Exp. 1 and 2, DeGMix demonstrates strong adaptability when integrating new tasks. For instance, in Exp. 2.3, after only 100 epochs of fine-tuning on the Boundary task, the model achieves an odsF of 76.6\% on NYUD-v2, confirming its rapid adaptation ability. Moreover, joint multi-task training (Exp. 2.4) still achieves superior overall performance compared to individual fine-tuning (Exp. 2.1–2.3), indicating that integrated optimization yields better generalization. These findings suggest that DeGMix enables efficient task expansion through fine-tuning, while full multi-task training remains optimal for overall consistency and generalization.

\noindent
\textbf{Evaluate new tasks using the trained model.}
As shown in Exp. 3 of Table~\ref{tab:diff_task_test_nyud_v2}, we adopt a leave-one-task-out strategy, where the model is trained on all tasks except one and then evaluated on the omitted task. The results (Exp. 3.1–3.4) reveal that excluding a task consistently weakens performance, as indicated by the orange numbers in Exp. 3. In particular, the \textit{S–N–B} model (Exp. 3.3) suffers notable degradation in depth, normal, and boundary estimation, while maintaining moderate segmentation performance. Conversely, the \textit{D–N–B} model (Exp. 3.4) struggles to capture semantic information, leading to performance drops across all tasks. These findings highlight the pivotal role of semantic segmentation and emphasize how task interactions shape multi-task learning effectiveness.

\begin{table*}[!t]
\begin{center}
\caption{Ablation studies and analysis on NYUD-v2 dataset using a Swin-T backbone. Deformable mixer (DM), task interaction (TI) block, and task query gating (TQG) block are the parts of our model. The Shared spatial gating (SSG) layer inserts into the TQG block. HR48 denotes HRNet48. MTB denotes the multi-task baseline. The notation ‘$\downarrow$’: lower is better. The notation ‘$\uparrow$’: Higher is better. The {w/} and {w/o} indicate {"with"} and "without", respectively.}
\scriptsize
	\centering
	\setlength{\tabcolsep}{4.pt}
	\subfloat[ \footnotesize{Ablation on components}]{
        
        \resizebox{0.480\textwidth}{!}{
		\begin{tabularx}{6.9cm}{c|c|c|c|c} 
		       \toprule[0.1em]
		       \multirow{2}*{Model}    & {SemSeg}   &Depth & Normal & Bound\\
                 & (mIoU)$\uparrow$   &(rmse)$\downarrow$   & (mErr)$\downarrow$  &(odsF)$\uparrow$\\
    	 	\hline
    	             MTB            &38.78    &0.6312   &21.05   &75.6\\
         \textit{w/} DM              &42.40 	  &0.6069   &20.83   &76.2\\
         \textit{w/} DM+SSG+TI       &45.24	  &0.5862   &20.55   &76.9\\
         DM+TI+\{TQG\textit{w/o}SSG\} &46.51   &0.5719   &20.42   &77.2\\
         DM+TI+TQG+SSG                &47.30   &0.5630   &20.14   &77.2\\
         
         \hline
         DeGMix Gain \textit{vs.} MTB       &+8.42    &+0.0652   &+0.9   &+1.6\\
			\bottomrule[0.1em]
		\end{tabularx} }
    } \label{tab:module}
    \hfill
    \setlength{\tabcolsep}{8.pt}
    \subfloat[\footnotesize{Ablation on the depths ($d$) of DM encoder}]{
        \resizebox{0.460\textwidth}{!}{
		\begin{tabularx}{6.4cm}{c|c|c|c|c} 
		\toprule[0.1em]
    		    \multirow{2}*{$d$} & {SemSeg}   &Depth & Normal & Bound\\ 
    		& (mIoU)$\uparrow$   &(rmse)$\downarrow$   & (mErr)$\downarrow$  &(odsF)$\uparrow$\\
    	 	\hline
          1  &46.96    &0.5717   &20.22   &77.1\\
          2  &47.30    &0.5630   &20.14   &77.2\\
          4  &48.12    &0.5601   &20.01   &77.3\\
          8  &48.01    &0.5521   &19.81   &77.2\\
			\bottomrule[0.1em]
		\end{tabularx}}
    }\label{tab:depth} 
    \hfill
    \setlength{\tabcolsep}{4.pt}
    \subfloat[\footnotesize{Ablation on scales}]{
        \resizebox{0.480\textwidth}{!}{
		\begin{tabularx}{7.2cm}{c|c|c|c|c} 
		  \toprule[0.1em]
    		  \multirow{2}*{Scale} & {SemSeg}   &Depth & Normal & Bound\\ 
    		  	   & (mIoU)$\uparrow$   &(rmse)$\downarrow$   & (mErr)$\downarrow$  &(odsF)$\uparrow$\\

    	 	\hline
             1/4                       &6.81     &1.0851   &31.65   &66.5\\
             1/4, 1/8                  &16.91	&0.8133   &27.06   &71.2\\
             1/4, 1/8, 1/16            &41.32    &0.6254   &21.18   &76.8\\
             1/4,1/8,1/16,1/32         &47.30    &0.5630   &20.14   &77.2\\
			\bottomrule[0.1em]
		\end{tabularx}}
    } \label{tab:scales}
    \hfill
    \subfloat[\footnotesize{Ablation on backbones}]{
        \setlength{\tabcolsep}{4.pt}
        \resizebox{0.460\textwidth}{!}{
	    \begin{tabularx}{6.9cm}{c|c|c|c|c}
		 \toprule[0.1em]
    		 \multirow{2}*{Backbone}  & {SemSeg}   &Depth & Normal & Bound \\ 
    		 & (mIoU)$\uparrow$   &(rmse)$\downarrow$   & (mErr)$\downarrow$  &(odsF)$\uparrow$\\
    		\hline
    	       HR48 baseline                 &41.96  &0.5543  &20.36   &77.6 \\
    	       HR48 \textbf{w/} ours         &46.87  &0.5409  &20.10   &77.8 \\
    	       Swin-B baseline               &51.44  &0.5813  &20.44   &78.0 \\
                  Swin-B+InvPT++\cite{invpt++}  &51.08  &0.5008  &19.28   &77.1 \\
                  Swin-B \textbf{w/} ours       &54.45  &0.5228  &19.33   &78.6 \\
        	\bottomrule[0.1em]
	    \end{tabularx}}
    } \label{tab:backbone}
    \hfill
\end{center}
\end{table*}

\subsection{Ablation Studies}
We carry out ablations using Swin-T on the NYUD-v2 dataset.
We ablate DeGMix to understand the individual contribution of each component and settings.
This analysis allows us to gain insights into the importance and effectiveness of each part in the overall performance of our method.

\noindent
\textbf{Ablation on modules.}
We analyze the influence of every key component and design choice in our full method.
All models are trained for 400 epochs on NYUD-v2.
The DeGMix model consists of three components: deformable mixer, task interaction, and task query gating blocks.
As shown in Table~\ref{tab:module}, we demonstrate the advantages of the deformable mixer, task interaction, and task query gating blocks.
After adding the proposed blocks to the multi-task baseline (MTB) network, the results improved persistently and notably.
Specifically, we observe that the task interaction block has more effect on the performance, and it is essential to interact with all the tasks for task interaction information.
This indicates that task interaction and task query gating blocks are essential to the task-aware gating transformer decoder.
From Figure~\ref{fig:vis_diff_module} and Table~\ref{tab:module} it can be observed that different components play a beneficial role.

\noindent
\textbf{Effectiveness of shared spatial gating (SSG).}
We further examine the role of the shared spatial gating (SSG) in the task query gating block. As shown in Table~\ref{tab:module}, integrating SSG into both the deformable mixer encoder and the task-aware gating Transformer decoder yields notable improvements—+8.52 mIoU (47.30 vs. 38.78) on SemSeg and a 0.0682 RMSE reduction (0.5630 vs. 0.6312). This confirms that SSG effectively enhances task-aware feature selection. Qualitative results in Figure~\ref{fig:vis_diff_module} also show that DeGMix surpasses the baseline MTB with DM (42.40 to 47.30 mIoU; 0.6069 to 0.5630 RMSE), validating the efficiency of shared gating across tasks.

\noindent
\textbf{Ablation on the depths $d$.}
As shown in Figure~\ref{fig:models}, the depth $d$ is the number of repetitions of the deformable mixer.
We add the $d$ to analyze the effect of the depth of the deformable mixer on the DeMT model.
In Table~\ref{tab:depth}, We vary the number of used deformable mixer depths ($e.g.,$ 1, 2, 4, 8) and compare their performances.
Comparing the first to last row in Table~\ref{tab:depth}, we observe the best performance when the depth is set to 4.
However, as increasing the depth, the parameters and FLOPs also become more extensive.
Practically, we choose a depth $d$ = 1 for all models in this paper.

\noindent
\textbf{Ablation on scales.}
We investigate the effect of multi-scale features on model performance.
As illustrated in Figure~\ref{fig:extractor}, the backbone generates features at four scales: 1/4, 1/8, 1/16, and 1/32.
Table~\ref{tab:scales} reports the performance under different scale combinations.
Model performance improves steadily with more feature scales. Using only the 1/4-scale features leads to a significant drop in accuracy, highlighting the necessity of multi-scale integration for richer semantic representation. By leveraging features from all four scales, our method effectively captures complementary semantic cues across tasks. Hence, we adopt four-scale features for all models to balance representation capacity and computational efficiency.

\noindent
\textbf{Ablation on backbones.}
Table~\ref{tab:backbone} compares DeGMix with different backbones (\textit{i.e.,}, HRNet48 and Swin-B).
As shown in Table~\ref{tab:backbone} and Figure~\ref{fig:params}, our DeGMix model with the Swin-B backbone achieves a notable improvement of +3.37 mIoU on the SemSeg task compared to the competitive InvPT++~\cite{invpt++}. 
In addition, we also observe the inspiring fact that using a larger transformer backbone can easily reach top-tier performance.
The different backbones are compared to demonstrate the generalization of our method.

\begin{table}[!t]
  \caption{ Ablation on shared spatial gating depth in task query gating. }
  \label{tab:ablation_gating_depth}
  \centering
\begin{tabular}{cllllllllll}
\toprule[0.1em]
 \multirow{2}*{$d$}     &{SemSeg}   &Depth &Normals  &Bound\\
      &(mIoU)$\uparrow$   &(rmse)$\downarrow$   &(mErr)$\downarrow$  &(odsF)$\uparrow$\\
\noalign{\smallskip}
\hline
\noalign{\smallskip}
1   &47.30  &0.5630 &20.14  &77.2 \\
2   &47.33	&0.5611	&20.12  &77.1 \\
4   &\textbf{47.55}	&0.5646	&20.12  &77.3 \\
6   &47.29	&0.5621	&20.05  &77.5 \\
8   &47.45	&\textbf{0.5587}	&\textbf{20.05}  &\textbf{77.6} \\
\bottomrule[0.1em]
\end{tabular}
\end{table}

\begin{figure*}[!t]
\centering 
\subfloat[\footnotesize{Semantic Segmentation}]{
\label{fig:seg}
\includegraphics[width=4.cm,height = 2.85cm]{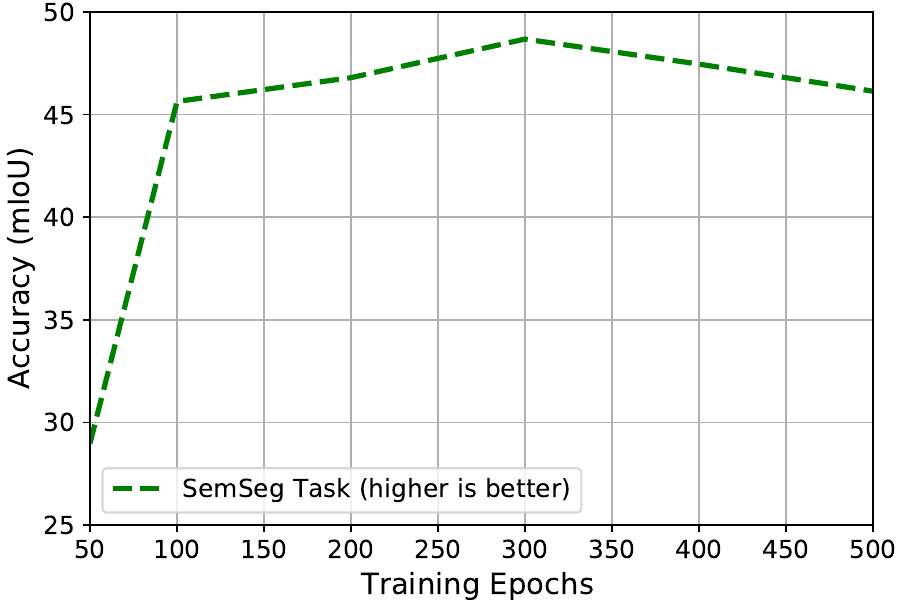}}
\subfloat[\footnotesize{Depth Estimation}]{
\label{fig:depth}
\includegraphics[width=4.cm,height = 2.85cm]{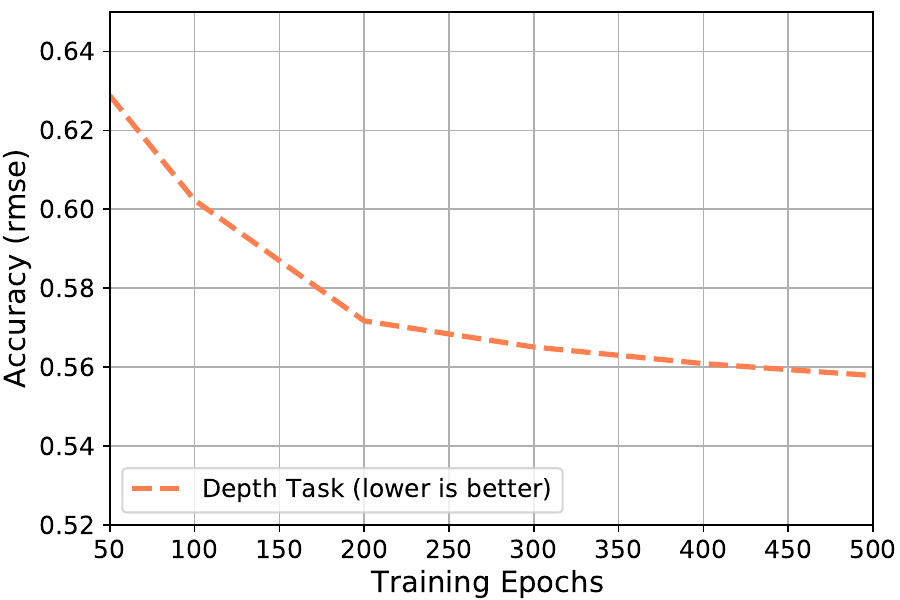}}
\subfloat[\footnotesize{Surface Normal Estimation}]{
\label{fig:norm}
\includegraphics[width=4.cm,height = 2.85cm]{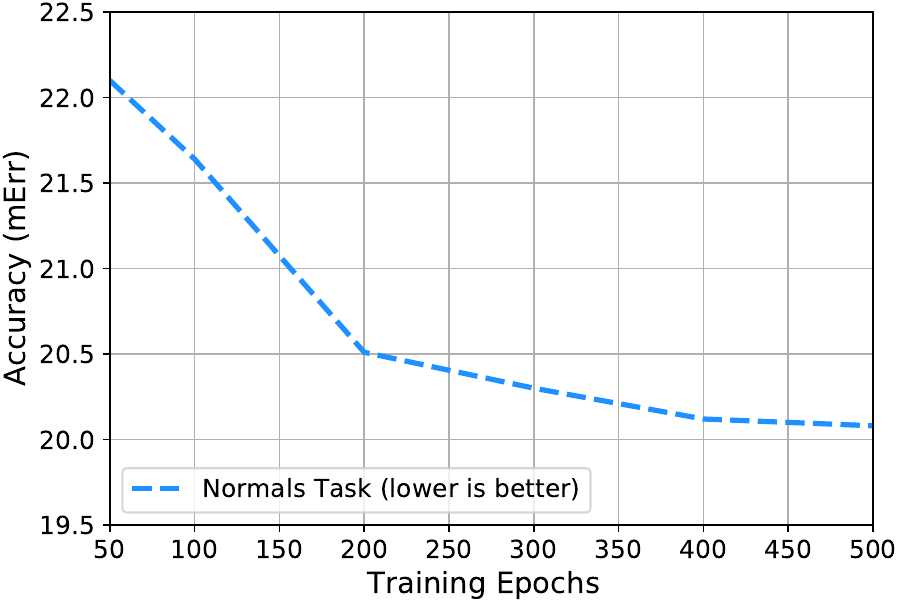}}
\subfloat[\footnotesize{Boundary Detection}]{
\label{fig:boun}
\includegraphics[width=4.cm,height = 2.85cm]{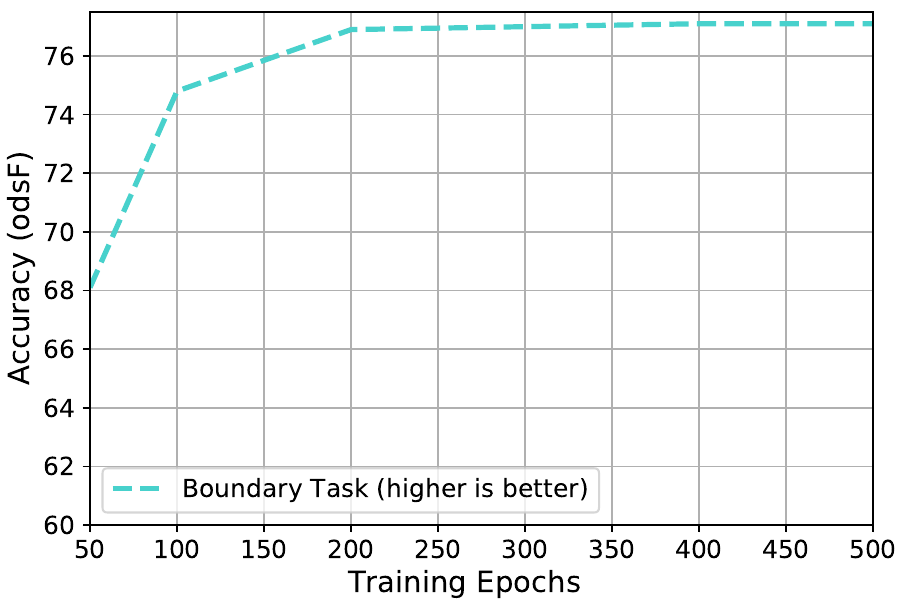}}
\caption{Influence of the number of epochs in multi-task training using Swin-T backbone on NYUD-v2. To evaluate the DeGMix performance of different epochs, we pick out models at the 50$^{th}$, 100$^{th}$, 200$^{th}$, 300$^{th}$, 400$^{th}$ and 500$^{th}$ epoch.}
\label{fig:epochs}
\end{figure*}

\noindent
\textbf{Ablation on the depths ($d$) of shared spatial gating in task query gating block.}
We analyze the impact of the shared spatial gating depth within the task query gating block for $d \in \{1, 2, 4, 6, 8\}$ (Table~\ref{tab:ablation_gating_depth}).
The results indicate that increasing the depth of the shared spatial gating provides marginal performance gains. However, these improvements come at the cost of a substantial increase in model parameters, which may limit the model's efficiency and scalability.
Using several depths shows a slight performance gain, and at a cost of significantly more parameters. 
Given this trade-off between accuracy and efficiency, we adopt $d=1$ as the default setting to maintain balanced performance and scalability.

\noindent
\textbf{Ablation on different epochs.}
To further evaluate the performance of DeGMix, we conduct an ablation study on the impact of training for different numbers of epochs, as shown in Figure~\ref{fig:epochs}.
We pick out models at the 50$^{th}$, 100$^{th}$, 200$^{th}$, 300$^{th}$, 400$^{th}$ and 500$^{th}$ epochs. 
Two key findings emerge: (1) performance improves with longer training until overfitting occurs beyond a certain point, and (2) the optimal epoch count varies across tasks.
For example, semantic segmentation peaks near 300 epochs, whereas other tasks continue to benefit from extended training.
These results emphasize the need for careful tuning of training epochs to balance convergence and generalization across tasks.

\begin{table*}[!ht]
\centering
\caption{Comparison of DeGMix-Tuning with existing fine-tuning methods on PASCAL-Context dataset. Results are reported using Swin-T/L backbones.
The “Single Inference for All Tasks” column indicates whether a model supports joint task execution.
‘$\downarrow$’ and ‘$\uparrow$’ denote that lower or higher values are better.
$\Delta_m$ denotes the average per-task performance drop (higher is better).}
\label{tab:finetune:pascal}
\begin{tabular}{llllllcc}
\toprule[0.1em]
 \multirow{2}*{Model}   & {SemSeg}   &PartSeg &Sal  & Normal  &\multirow{2}*{$\Delta_m[\%]$$\uparrow$}  &Trainable  &Single Inference\\
     & (mIoU)$\uparrow$  & (mIoU)$\uparrow$  &(mIoU)$\uparrow$  &(mErr)$\downarrow$& &Parameters (M) &For All Tasks\\
\noalign{\smallskip}
\hline
\noalign{\smallskip}
single task baseline            &67.21 &61.93 &62.35 &17.97 &0 &112.62    &\ding{55}\\
MTL-Tuning Decoder Only         &65.09 &53.48 &57.46 &20.69 &-9.95 &1.94  &\ding{51}\\
MTL - Full Fine Tuning          &67.56 &60.24 &65.21 &16.64 &+2.23 &30.06 &\ding{51}\\
\hline
Adapte~\cite{he2022adapter}     &69.21 &57.38 &61.28 &18.83 &-2.71 &11.24 &\ding{55}\\
Bitfit~\cite{bitfit2022}        &68.57 &55.99 &60.64 &19.42 &-4.60 &2.85 &\ding{55}\\
VPT-shallow~\cite{VPTshallow2022}&62.96 &52.27 &58.31 &20.90 &-11.18 &2.57 &\ding{55}\\
VPT-deep~\cite{VPTshallow2022} &64.35 &52.54 &58.15 &21.07 &-10.85 &3.43 &\ding{55}\\
LoRA~\cite{hu2022lora}         &70.12 &57.73 &61.90 &18.96 &-2.17 &2.87 &\ding{55}\\
Polyhistor~\cite{Polyhistor22}  &70.87 &59.54 &65.47 &17.47 &+2.34 &8.96 &\ding{55}\\
MTLoRA~\cite{MTLoRA24}          &68.19 &58.99 &64.48 &17.03 &+1.35 &4.95 &\ding{51}\\
DeGMix-Tuning (Ours)      &67.37 &56.87 &64.12 &15.67 &+1.96 &2.2  &\ding{51}\\
\hdashline
DITASK+Swin-L                   &76.23 &67.53 &64.07 &16.90 &+7.79 &7.13 &\ding{51}\\
DeGMix-Tuning+Swin-L (Ours)     &78.14 &64.44 &64.20 &16.30 &+8.01 &3.2  &\ding{51}\\
\bottomrule[0.1em]
  \end{tabular}
\end{table*}

\subsection{Parameter Efficient Training}
\noindent
{Parameter Efficient Training on PASCAL-Context.}
For a fair comparison, we evaluate DeGMix-Tuning with Swin-T and Swin-L backbones (Table~\ref{tab:finetune:pascal}). With only 2.2M trainable parameters, DeGMix-Tuning achieves a +1.96 $\Delta_m$ improvement, demonstrating strong efficiency–performance trade-offs. Although HyperFormer (+2.64), Polyhistor (+2.34), and MTL-Full Fine-Tuning (+2.23) yield slightly higher $\Delta_m$, they require 33$\times$, 4$\times$, and 13.7$\times$ more parameters, respectively. When scaled to Swin-L, our method attains the best overall gain (+8.01 $\Delta_m$) using just 3.2M parameters, surpassing DITASK~\cite{ditask2025} (+7.79 $\Delta_m$, 7.13M params). Overall, DeGMix-Tuning achieveshigh parameter efficiency by effectively transferring multi-task knowledge with minimal additional cost.

\begin{figure}[!t]
\centering
  \includegraphics[width=0.48\textwidth]{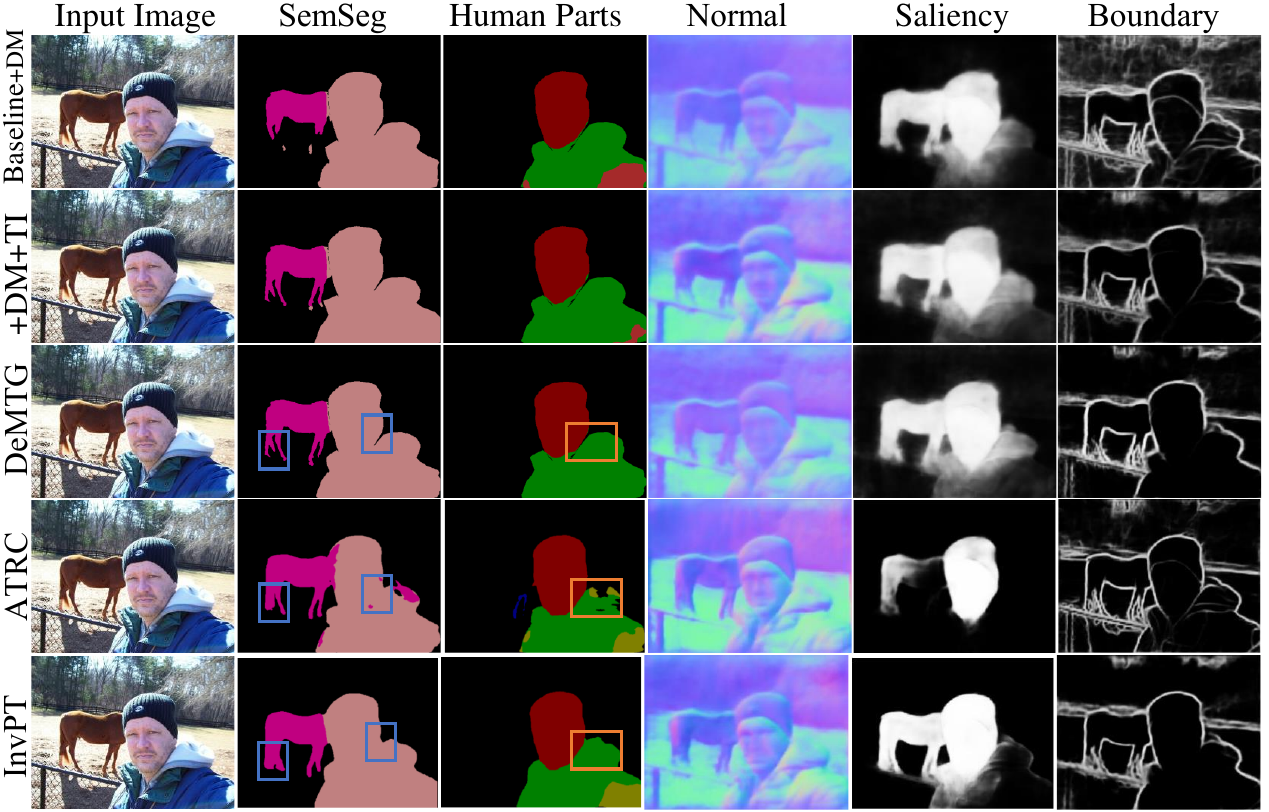}
  \caption{ Visual results on 5-task PASCAL-Context dataset.
  Qualitative analysis of the components according to Table~\ref{tab:module}.
  We also select the visualization results of ATRC~\cite{atrc_2021} and InvPT~\cite{InvPT_2022} as a comparison. Note that they are marked in blue and orange square boxes. The qualitative comparison clearly proves the efficacy of our model in capturing the objects' subtle contextual details.
}
  \label{fig:vis_diff_module}
\end{figure}

\noindent
\textbf{Qualitative Comparison.}
Figure~\ref{fig:vis_diff_module} illustrates the visual impact of each component.
While the Baseline + DM (deformable mixer) struggles to recover certain objects, DeGMix produces clearer and more complete predictions across multiple tasks.
Compared with ATRC~\cite{atrc_2021} and InvPT~\cite{InvPT_2022}, DeGMix yields markedly sharper boundaries and more accurate semantic and part segmentation, especially on PASCAL-Context.
Consistent with quantitative results, these visualizations confirm the superior task consistency and robustness of DeGMix.

\begin{figure}[!t]
\centering
  \includegraphics[width=0.48\textwidth]{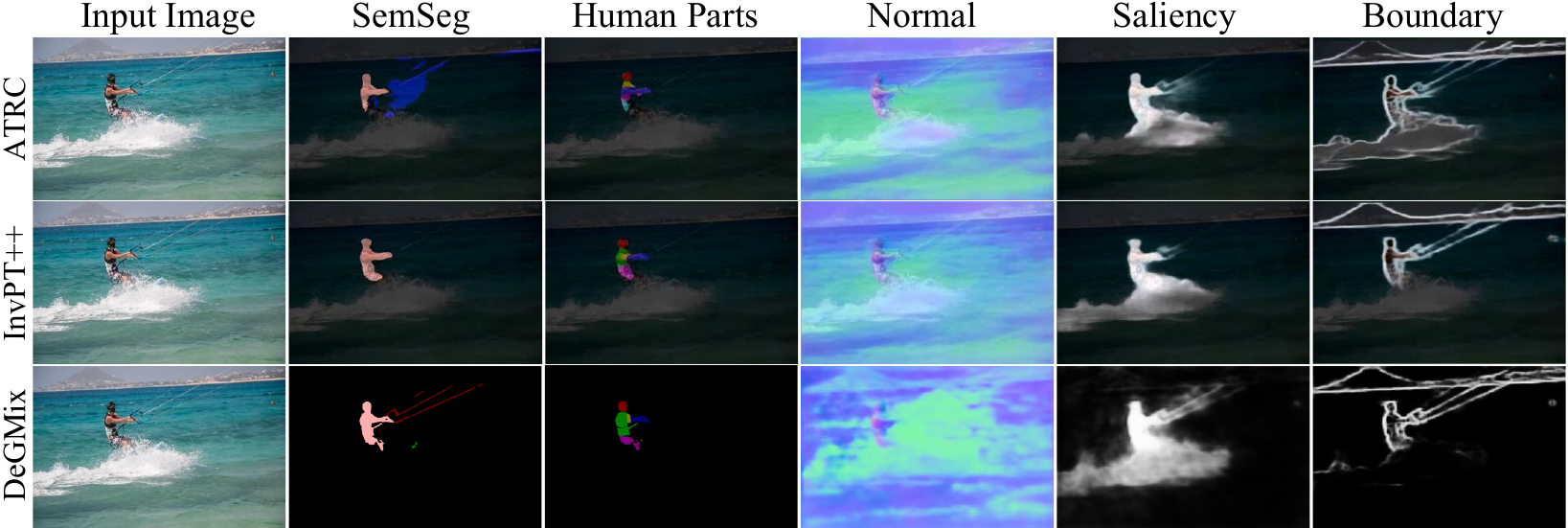}
  \caption{ Visual results of the generated 5-task predictions on DAVIS dataset.
  We also select the visualization results of ATRC~\cite{atrc_2021} and InvPT++~\cite{invpt++} as a comparison. The model is trained with PASCAL-Context dataset. The qualitative comparison clearly proves the efficacy of our model in capturing detailed segmentation of subtle objects (\textit{e.g.,} the flying lines and control bar). 
}
  \label{fig:vis_diff_davis}
\end{figure}

\noindent
\textbf{Generalization Performance}.
Figure~\ref{fig:vis_diff_davis} presents qualitative results on the DAVIS dataset, highlighting the strong generalization ability of DeGMix across multiple tasks. Compared with ATRC~\cite{atrc_2021} and InvPT++~\cite{invpt++}, DeGMix produces more precise and coherent predictions, with notably sharper and more detailed segmentation boundaries in complex scenes. The results show that DeGMix generalizes effectively to diverse scenarios and consistently outperforms competing methods.

\section{Conclusion}
\label{sec:conclusion}
In this work, we introduce DeGMix, a simple and effective approach that unifies deformable CNNs and query-based Transformers with spatial gating for multi-task dense prediction.
A key innovation of our method is the use of deformed features, generated by the deformable mixer encoder, as task queries within the task-aware gating transformer decoder. The model employs query-based attention and a gating mechanism to generate dynamically selected task awareness features, enabling efficient learning of multiple related tasks.
Extensive experiments on NYUD-v2, PASCAL-Context, and Cityscapes demonstrate that DeGMix achieves superior accuracy with substantially fewer trainable parameters, validating its efficiency and generality.
DeGMix surpasses the accuracy of a fully fine-tuned MTL model while requiring substantially fewer trainable parameters.

\noindent
\textbf{Limitations and future work.}
While DeGMix achieves strong multi-task performance, its parameter and computational costs scale with task number and network depth. Future work will explore more efficient designs, such as fine-tuning task-specific subsets to preserve accuracy with lower complexity.

\bibliographystyle{IEEEtran}
\bibliography{ref23}   

\newpage

\vfill
\end{document}